\documentclass[lettersize,journal]{IEEEtran}
\usepackage{amsmath,amsfonts}
\usepackage{algorithmic}
\usepackage{algorithm}
\usepackage{array}
\usepackage[caption=false,font=normalsize,labelfont=sf,textfont=sf]{subfig}
\usepackage{textcomp}
\usepackage{stfloats}
\usepackage{url}
\usepackage{verbatim}
\usepackage{graphicx}
\usepackage{cite}
\usepackage{bm}
\usepackage{amssymb}
\usepackage{multirow}
\usepackage{booktabs}
\usepackage{fontawesome5}
\usepackage[table]{xcolor}

\usepackage[colorlinks, linkcolor=blue, anchorcolor=blue, citecolor=blue]{hyperref}
\usepackage{orcidlink}

\begin{document}

\title{Analytic Drift Resister for Non-Exemplar Continual Graph Learning}

\author{Lei Song\orcidlink{0000-0003-2044-3311}, Shihan Guan\orcidlink{0000-0003-3563-5308}, Youyong Kong
\thanks{\faBullhorn~\textbf{Note: This work has been submitted to the IEEE for possible publication. Copyright may be transferred without notice, after which this version may no longer be accessible.}}

\thanks{Lei Song, Shihan Guan, and Youyong Kong are with the School of Computer Science and Engineering, Southeast University, Nanjnig 211189, China (e-mail: 230238577@seu.edu.cn; 220225771@seu.edu.cn; kongyouyong@seu.edu.cn).}
}

\markboth{Journal of \LaTeX\ Class Files,~Vol.~14, No.~8, August~2021}%
{Shell \MakeLowercase{\textit{et al.}}: A Sample Article Using IEEEtran.cls for IEEE Journals}

\IEEEpubid{}

\maketitle

\begin{abstract}
Non-Exemplar Continual Graph Learning (NECGL) seeks to eliminate the privacy risks intrinsic to rehearsal-based paradigms by retaining solely class-level prototype representations rather than raw graph examples for mitigating catastrophic forgetting. However, this design choice inevitably precipitates feature drift. As a nascent alternative, Analytic Continual Learning (ACL) capitalizes on the intrinsic generalization properties of frozen pre-trained models to bolster continual learning performance. Nonetheless, a key drawback resides in the pronounced attenuation of model plasticity. To surmount these challenges, we propose Analytic Drift Resister (ADR), a novel and theoretically grounded NECGL framework. ADR exploits iterative backpropagation to break free from the frozen pre-trained constraint, adapting to evolving task graph distributions and fortifying model plasticity. Since parameter updates trigger feature drift, we further propose Hierarchical Analytic Merging (HAM), performing layer-wise merging of linear transformations in Graph Neural Networks (GNNs) via ridge regression, thereby ensuring absolute resistance to feature drift. On this basis, Analytic Classifier Reconstruction (ACR) enables theoretically zero-forgetting class-incremental learning. Empirical evaluation on four node classification benchmarks demonstrates that ADR maintains strong competitiveness against existing state-of-the-art methods.
\end{abstract}


\section{Introduction}

\begin{figure*}[t]
  \begin{center}
    \centerline{\includegraphics[width=\textwidth, trim=0 10 0 0, clip]{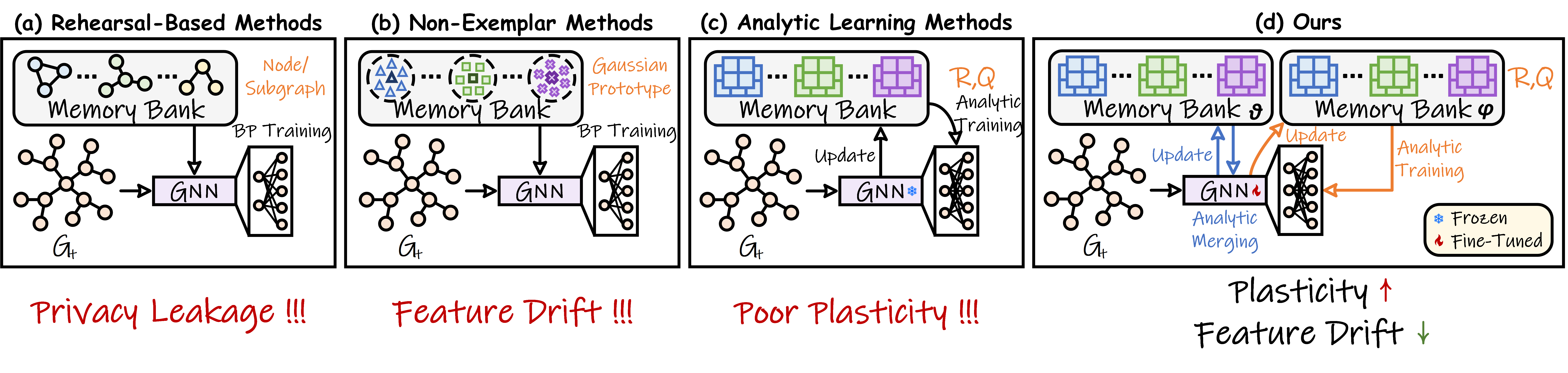}}
    \caption{
    A schematic overview of different Continual Graph Learning paradigms.
    }
    \label{fig1}
  \end{center}
\end{figure*}

\IEEEPARstart{G}{raph} Neural Networks (GNNs) have manifested exceptional efficacy across a broad spectrum of graph data analysis applications, including citation network analysis~\cite{kipf2016semi, hamilton2017inductive}, recommender systems~\cite{he2020lightgcn, yang2024decoupled}, and neurological disorder diagnosis~\cite{yang2023deep, li2025topology}, among others. Most existing studies revolve around static task scenarios, where GNNs are trained on pre-collected, task-specific data and remain immutable thereafter. However, the dynamic real world incessantly yields new data, while human demands progressively evolve through sustained environmental interactions. This calls for continual parameter adaptation in GNNs to maintain long-term applicability. A na\"{i}ve solution is to incrementally fine-tune the model on incoming task data, yet this typically incurs substantial performance degradation on prior tasks owing to pronounced task distribution shifts, a phenomenon referred to as \emph{catastrophic forgetting}. Accordingly, Continual Graph Learning (CGL) seeks to endow GNNs with sustained adaptability to new tasks while safeguarding prior knowledge.

Rehearsal-based paradigms~\cite{zhou2021overcoming, zhang2022sparsified, arani2022learning} constitute a potent strategy against catastrophic forgetting, capitalizing on a compact memory bank of historical examples (e.g., old-class nodes or subgraphs) to retain sensitivity to previously acquired knowledge during adaptation to new tasks. Nonetheless, a critical drawback lies in violating data privacy, barring use in sensitive contexts. To avert this risk, Non-Exemplar Continual Graph Learning (NECGL)~\cite{ren2023incremental, wang2023non, magistri2024elastic} orchestrates class prototypes as proxies for historical task distributions to steer the model. This paradigm derives class-wise means and covariances (i.e., gaussian prototypes) via the trained encoder at each task’s completion, archiving them in the memory bank for later replay. However, continual updates to the model parameters cause cached prototypes to progressively diverge from their true positions, resulting in the notorious feature drift. Although post-hoc drift compensation can mitigate this issue to some extent, the approximation remains vulnerable to task distribution shifts, especially in graph tasks where variations in node features and topology amplify the error. Analytic Continual Learning (ACL)~\cite{zhuang2022acil, zhuang2023gkeal, zhuang2024gacl} has recently arisen as a promising alternative. Such approaches capitalize on the inherent generalization of frozen pre-trained models to extract task-relevant features and determine linear classifier parameters via recursive linear regression. Exclusive retention of the autocorrelation and cross-correlation feature matrices, denoted as $R$ and $Q$, renders private information irretrievable, thereby eliminating associated privacy risks. Fixing the encoder parameters effectively forestalls feature drift, yet concurrently compromises model plasticity, impeding adaptation to incremental tasks. Recent studies~\cite{zhuang2024ds, he2025semantic} have identified this issue but remain confined to the frozen-parameter or post-hoc compensation paradigms. Fig.~\ref{fig1} illustrates the three CGL paradigms in a schematic form.

In this paper, we present Analytic Drift Resister (ADR), a theoretically grounded NECGL framework that improves model plasticity while eliminating feature drift, and inherits the privacy-preserving properties of ACL. ADR relaxes the frozen-parameter constraint in ACL and updates GNNs via backpropagation (BP) on each incremental task graph, enabling close alignment with the evolving task distribution. However, parameter updates may induce feature drift, exacerbating catastrophic forgetting. Inspired by model merging techniques~\cite{matena2022merging, yadav2023ties, marczak2024magmax}, we further propose Hierarchical Analytic Merging (HAM), which performs layer-wise parameter merging of GNNs via solving ridge regression~\cite{hoerl1970ridge}. From a theoretical standpoint, incremental HAM is equivalent to the joint training objective, conferring absolute resistance to feature drift. On this basis, Analytic Classifier Reconstruction (ACR) infers linear classifier weights from the merged encoder’s outputs via the closed-form solution of ridge regression. Formally equivalent to the joint training objective, ACR enables zero-forgetting class-incremental learning. Accordingly, ADR embodies a dual theoretical advantage over existing approaches. The main contributions are summarized as follows:
\begin{itemize}
  \item We present ADR, a theoretically grounded NECGL framework endowed with fortified model plasticity, absolute resistance to feature drift, and intrinsic privacy preservation.
  \item We dismantle the frozen pre-trained parameter constraint, freeing GNNs for adaptation to evolving incremental task graph distributions, markedly bolstering model plasticity.
  \item To overcome feature drift arising from model updates, we propose HAM, which consolidates GNNs’ parameters layer-wise, theoretically conferring absolute resistance. Drawing upon the merged encoder, ACR theoretically formalizes zero-forgetting class-incremental learning via the closed-form solution of ridge regression.
  \item Extensive empirical evidence on four node classification benchmark datasets demonstrates that ADR substantially enhances model plasticity, achieving competitive performance compared to existing state-of-the-art (SOTA) methods.
\end{itemize}

\section{Related Works}

\subsection{Continual Graph Learning}

In this section, we examine existing CGL literature pertinent to our work, encompassing rehearsal-based, Non-Exemplar, and Analytic Continual Learning methods.

\textbf{Rehearsal-based methods} safeguard model acuity to previously acquired knowledge while adapting to novel tasks by continuously revisiting historical exemplars cached in a compact memory bank. This line of research has largely been devoted to the improvement of exemplar sampling protocols and the principled formulation of compact exemplar storage schemes. For instance, \cite{zhou2021overcoming} framed individual nodes as storage units and devised three experience selection strategies to retain the most representative examples from the entire task graph. Given the risk of losing discriminative information from discarded neighborhood topology, \cite{zhang2022sparsified} sampled central subgraphs and sparsified them before memory caching, substantially reducing storage overhead. \cite{liu2023cat, niu2024graph} condensed the task graph into an informative synthetic replay graph, followed by unbiased learning on its class-balanced counterpart. Although these methods demonstrate remarkable efficacy, retaining raw examples often poses significant data privacy risks. While graph condensation~\cite{jin2021graph, liu2022graph} can alleviate this issue to some extent, training synthetic replay graphs incurs considerable computational cost.

\textbf{Non-Exemplar methods} instantiate gaussian prototypes as task proxies, obviating the privacy vulnerabilities inherent in native exemplar retention. More concretely, when training on a new task, old-class embeddings are stochastically drawn from cached gaussian prototypes and orchestrated via prototype replay (PR)~\cite{zhu2021prototype, ren2023incremental, li2024fcs} to alleviate catastrophic forgetting, while knowledge distillation further consolidates prior memory. However, a critical issue stems from the incongruence of old-class prototypes with the incrementally updated embedding space, giving rise to feature drift. While obsolete old-class prototypes can be corrected via post-hoc or online drift compensation~\cite{wang2023non, gomez2024exemplar, cheng2024efficient, magistri2024elastic}, the lack of reusable historical exemplars restricts drift estimation to the current task. Significant cross-task distributional shifts can amplify drift estimation errors, compromising prior task memory.

\textbf{Analytic Continual Learning methods} recast class-incremental learning as a linear regression problem, yielding a closed-form solution via Least Squares (LS)~\cite{guo2004pseudoinverse, zhuang2021blockwise}. In particular, most existing studies~\cite{zhuang2022acil, zhuang2023gkeal, zhuang2024gacl, he2025real} capitalize on the generalization of a pre-trained encoder to extract incremental-task features, followed by recursive computation of class-incremental classifier weights via LS. However, this paradigm suffers from impaired model plasticity under significant distributional discrepancies between pre-training and incremental tasks. \cite{zhuang2024ds, yue2024mmal} proposed compensation branches to address the underfitting dilemma, yet this strategy remains constrained by frozen pre-trained weights. Although \cite{he2025semantic} released the frozen parameter constraint to boost model plasticity, feature drift persists as a critical impediment to the approach.

\subsection{Model Merging}

Model merging techniques~\cite{jin2022dataless} aim to consolidate the weights of independently fine-tuned models with shared architectures, retaining their performance across multiple tasks. Existing studies primarily focus on devising principled merging strategies to avert post-merging performance degradation. \cite{choshen2022fusing, wortsman2022model} proposed directly averaging the weights of multiple individual models. \cite{matena2022merging} exploited the Fisher information matrix to orchestrate parameter merging in accordance with their relative importance. \cite{yadav2023ties, marczak2024magmax} conducted a systematic analysis of inter-model parameter interference and proposed a more rational merging paradigm based on task vector~\cite{ilharco2022editing}. In this paper, we present a novel model merging strategy to consolidate prior task-specific models, achieving absolute resistance to feature drift.

\begin{figure*}[t]
  \begin{center}
    \centerline{\includegraphics[width=0.8\textwidth, trim=0 0 0 0, clip]{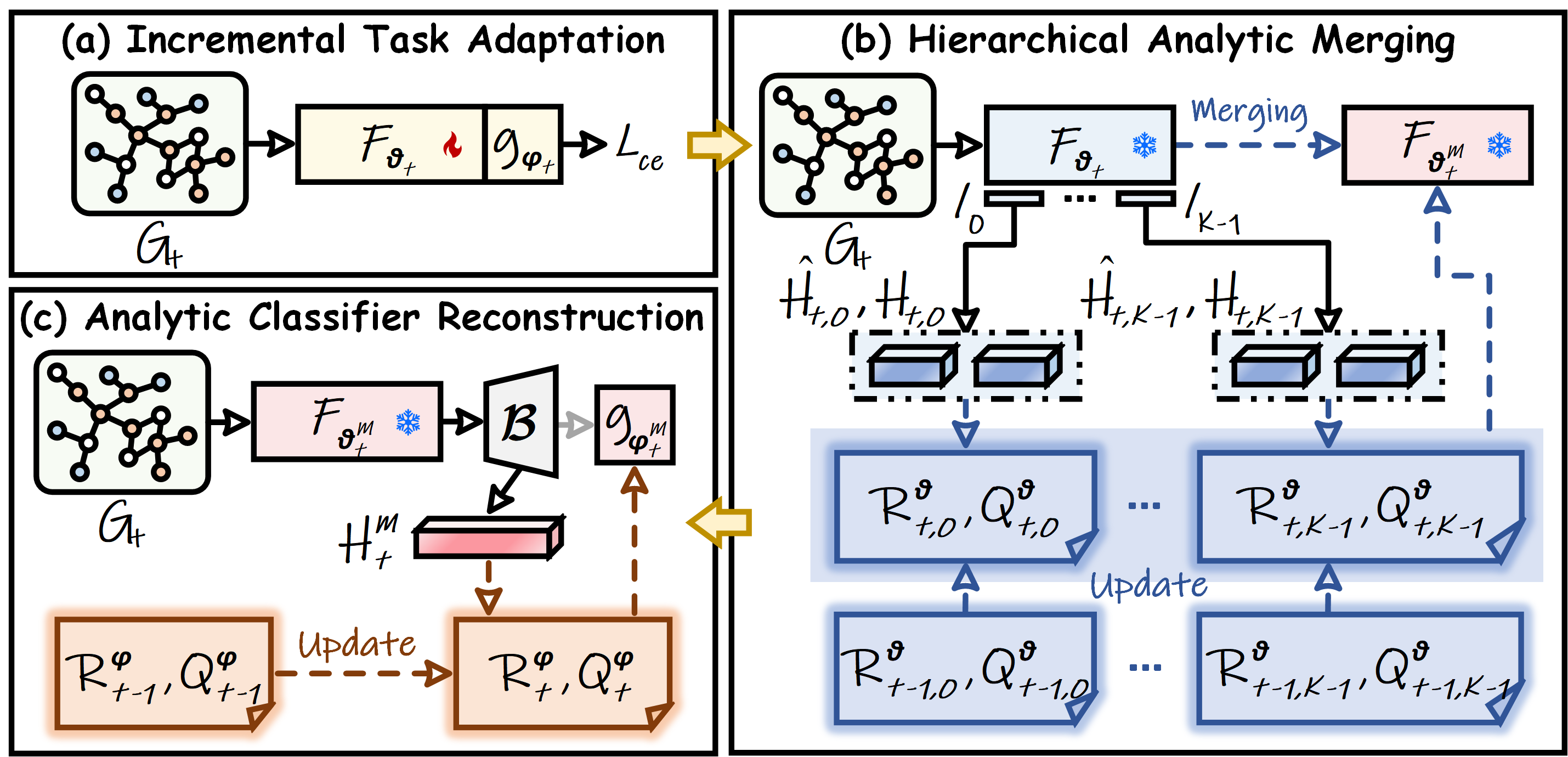}}
    \caption{
    The overall pipeline of the proposed ADR. (a) Upon the arrival of a new task $\mathcal{T}_t$, the model adapts freely to the corresponding graph distribution via iterative BP without any imposed regularization. (b) HAM performs analytic merging of the layer-wise linear transformations of all historical task encoders, unifying their latent representation spaces. (c) ACR exploits the merged encoder to derive linear classifier weights, enabling robust class-incremental learning.
    }
    \label{fig2}
  \end{center}
\end{figure*}

\section{Methodology}

Fig.~\ref{fig2} illustrates the overall pipeline of our ADR. The framework comprises three fundamental stages: Incremental Task Adaptation, Hierarchical Analytic Merging, and Analytic Classifier Reconstruction. We next systematically present each component in detail.

\subsection{Incremental Task Adaptation}

Before delving into the technical details of ADR, we provide a rigorous problem formulation. In the class-incremental learning scenario, GNNs are mandated to undergo iterative updates over a chronologically ordered task stream $\mathcal{T}=\{\mathcal{T}_0, \mathcal{T}_1, ..., \mathcal{T}_{N-1}\}(\left|\mathcal{T}\right|=N)$, with each task $\mathcal{T}_{t< N}=\{\mathcal{G}_t, \mathcal{Y}_t\}$ defined as a semi-supervised node classification problem. The task graph $\mathcal{G}_t$ comprises the node set $\mathcal{V}_t$ and edge set $\mathcal{E}_t$, denoted as $\mathcal{G}_t=\{\mathcal{V}_t, \mathcal{E}_t\}$. Moreover, $\mathcal{V}_t$ and $\mathcal{E}_t$ can be instantiated as a node feature matrix $\mathbf{X}_t\in \mathbb{R}^{\left|\mathcal{V}_t\right|\times d}$ and a binary adjacency matrix $\mathbf{A}_t\in\mathbb{R}^{\left|\mathcal{V}_t\right| \times \left|\mathcal{V}_t\right|}$, whose entries encode the presence (1) or absence (0) of edges. The label set is given by $\mathcal{Y}_t=\{y_t^{0}, y_t^{1}, ..., y_t^{c_{t}-1}\}(\left|\mathcal{Y}_t\right|=c_t)$, with label spaces being mutually exclusive across tasks, i.e., $\mathcal{Y}_i \cap \mathcal{Y}_j=\varnothing$ for all $i \neq j$. In this paper, we adhere to the strict Non-Exemplar assumption under which, at time step $t$, data from preceding tasks $\mathcal{T}_{0:t-1}$ are entirely inaccessible. After training, the GNNs are tasked with classifying all classes observed in the cumulative task stream $\mathcal{T}_{0:t}$.

While existing ACL methods capitalize on the generalization prowess of pre-trained encoders to attenuate catastrophic forgetting and achieve tangible performance gains, the rigid exclusion of parameter updates fundamentally erodes model plasticity, curtailing thorough assimilation of new knowledge. In graph-based tasks, pronounced heterogeneity in node attributes and topological structures can critically impede the transferability of pre-trained knowledge. We propose to abolish this undue constraint, enabling GNNs to engage in free exploration on new task graphs. Consider a model comprising a graph encoder $\mathcal{F}_{\theta_t}(\cdot)$ and a linear classifier $g_{\phi_t}(\cdot)$, whose weights are parameterized by $\theta_t$ and $\phi_t$. Given the current task graph data $\mathbf{X}_t$ and $\mathbf{A}_t$, the model derives node embeddings via iterative message passing as follows:
\begin{equation}
\label{eq1}
\mathbf{H}_{t,k}=\mathrm{GNN}_{t,k}(\mathbf{H}_{t,k-1}, \mathbf{A}_t),\quad k<K,
\end{equation}
where $\mathbf{H}_{t,-1}=\mathbf{X}_t$, and $\mathrm{GNN}_{t,k}$ denotes the $k$-th layer of the model. In task $\mathcal{T}_t$, we minimize the task-specific cross-entropy loss for incremental task adaptation. The objective function is expressed as follows:
\begin{equation}
\label{eq2}
    \mathcal{L}_{ce}=\mathbb{E}_{(x_t,y_t)\sim\mathcal{T}_t} \left[-\log p_{\Theta_t}(y_t|x_t)\right],
\end{equation}
where $\Theta_t=\{\theta_t,\phi_t\}$ constitutes the full set of model parameters, with $p_{\Theta_t}(y_t|x_t)$ representing the probability distribution over $y_t$.

Notably, rather than imposing logit- or feature-level distillation~\cite{li2017learning, ren2023incremental}, we prioritize maximizing model plasticity for complete adaptation to novel tasks.

\subsection{Hierarchical Analytic Merging}

A key driver of feature drift in NECGL is the discord between evolving encoder parameters and static cached features. Incremental Task Adaptation likewise engenders this effect, leading to significant performance decline. Inspired by recent advances in model merging, we propose HAM to entirely circumvent this issue.

At time step $t$, given a collection of encoder parameters $\{\theta_i\}_{i=0}^{t}$ learned from preceding tasks $\mathcal{T}_{0:t}$, we seek to consolidate task-specific knowledge from all tasks into a unified, architecture-consistent encoder $\mathcal{F}_{\theta_{t}^{M}}(\cdot)$, parameterized by $\theta_{t}^{M}$. The merged encoder is devised to faithfully reproduce the task-specific behaviors exhibited by each individual encoder, thus fully eliminating feature drift. To this end, we establish the following optimization problem:
\begin{equation}
\label{eq3}
{\theta_{t}^{M}}^*=\mathop{\arg\!\min}_{\theta_t^M}\sum_{i=0}^{t}\mathbb{E}_{(x_i,y_i)\sim\mathcal{T}_{i}}\left[\ell_{d}(h_{i,K-1}, h_{t,K-1}^M)\right],
\end{equation}
where $h_{i,K-1}\!\in\!\mathbf{H}_{i,K-1}$ and $h_{t,K-1}^M\!\in\!\mathbf{H}_{t,K-1}^M$ denote the node embeddings from the $K$-layer encoder $\mathcal{F}_{\theta_{i}}(\cdot)$ trained on task $\mathcal{T}_i$ and the merged encoder $\mathcal{F}_{\theta_{t}^M}(\cdot)$ with an identical architecture, respectively, and $\ell_d$ is a distance-based loss function. The optimal merged weights ${\theta_t^{M}}^*$ can be derived via iterative BP to fulfill the prescribed target. Considering that the parameters of GNNs are primarily governed by the linear transformations in each layer, we further reformulate Eq.~\eqref{eq3} as follows:
\begin{equation}
\label{eq4}
{\theta_{t}^{M}}^*=\mathop{\arg\!\min}_{\theta_t^M}\sum_{i=0}^{t}\sum_{k=0}^{K-1}\mathbb{E}_{(x_i,y_i)\sim\mathcal{T}_{i}}\left[\ell_{d}(h_{i,k}, h_{t,k}^M)\right].
\end{equation}
Given the target node $v_o$, we have:
\begin{equation}
\label{eq5}
h_{i,k}^o=\sigma\left(\sum_{j\in\mathcal{N}(o)}\omega_{oj} W_{\theta_{i,k}}^\top h_{i,k-1}^j\right),
\end{equation}
where $\mathcal{N}(o)$ denotes the neighborhood of $v_o$ including itself, $\omega_{oj}$ is the weighting coefficient for message passing, $W_{\theta_{i,k}}\!\in\!\mathbb{R}^{d_{k}^{in}\times d_{k}^{out}}$ represents the linear transformation matrix of the $k$-th layer in $\mathcal{F}_{\theta_i}(\cdot)$, and $\sigma(\cdot)$ defines a nonlinear activation function. Analogously, $h_{t,k}^M$ is also characterized. However, a key obstacle in solving Eq.~\eqref{eq4} is the inaccessibility of historical data in the Non-Exemplar scenario. Inspired by the blockwise recursive Moore-Penrose inverse (BRMP)~\cite{zhuang2021blockwise}, we formulate a recursive closed-form solution that enables analytic computation of $\{W_{\theta_{t,k}^M}^*\}_{k=0}^{K-1}$ via a single forward pass. Let $\hat{h}_{i,k}^o=\sum_{j\in\mathcal{N}(o)}\omega_{oj} h_{i,k-1}^j$. For clarity, we overload the notation as $h_{i,k}^o=W_{\theta_{i,k}}\hat{h}_{i,k}^o$. On this basis, we formulate the ridge regression objective for solving $W_{\theta_{t,k}^M}^*$ as follows:
\begin{equation}
\label{eq6}
\mathop{\arg\!\min}_{W_{\theta_{t,k}^M}}\left\|\left[
\begin{array}{c}
\mathbf{H}_{0,k} \\
\mathbf{H}_{1,k} \\
\vdots \\
\mathbf{H}_{t,k}
\end{array}
\right] - 
\left[
\begin{array}{c}
\hat{\mathbf{H}}_{0,k} \\
\hat{\mathbf{H}}_{1,k} \\
\vdots \\
\hat{\mathbf{H}}_{t,k}
\end{array}
\right]W_{\theta_{t,k}^M}
\right\|_{F}^2 + \gamma\left\|W_{\theta_{t,k}^M}\right\|_F^2,
\end{equation}
where $\left\|\cdot\right\|_F^2$ represents the Frobenius norm, and $\gamma$ governs the regularization term for preventing ill-conditioning. Owing to convexity, the optimal solution is obtained when the gradient with respect to $W_{\theta_{t,k}^M}$ is zero.
\begin{equation}
\label{eq7}
W_{\theta_{t,k}^M}^*=\left(\sum_{i=0}^{t}\hat{\mathbf{H}}_{i,k}^\top \hat{\mathbf{H}}_{i,k}+\gamma \mathbf{I}\right)^{-1}\sum_{i=0}^{t}\hat{\mathbf{H}}_{i,k}^\top \mathbf{H}_{i,k},
\end{equation}
where $\mathbf{I}\in\mathbb{R}^{d_k^{in}\times d_k^{in}}$ denotes the identity matrix. This equation reveals that retaining only the accumulated autocorrelation feature matrix $R_{t,k}^\theta=\sum_{i=0}^{t}\hat{\mathbf{H}}_{i,k}^\top \hat{\mathbf{H}}_{i,k}$ and cross-correlation feature matrix $Q_{t,k}^\theta=\sum_{i=0}^{t}\hat{\mathbf{H}}_{i,k}^\top \mathbf{H}_{i,k}$ suffices to consolidate linear transformation parameters across the entire task history, thus unifying the latent space to eradicate feature drift. Analogously, this holds for each layer, with the encoder-side memory bank denoted by $\{(R_{t,k}^\theta, Q_{t,k}^\theta)\}_{k=0}^{K-1}$. For a $K$-layer encoder, the memory bank sustains a constant storage footprint of $\mathcal{O}\left(\sum_{k=0}^{K-1}{d_k^{in}}^2+d_k^{in}d_k^{out}\right)$. We present a rigorous mathematical derivation of Eq.~\eqref{eq7} below.

\emph{proof}. Let $\bm{\mathcal{H}}_{t,k}=\left[
\begin{array}{c}
\mathbf{H}_{0,k} \\
\mathbf{H}_{1,k} \\
\vdots \\
\mathbf{H}_{t,k}
\end{array}
\right]$ and $\hat{\bm{\mathcal{H}}}_{t,k}=\left[
\begin{array}{c}
\hat{\mathbf{H}}_{0,k} \\
\hat{\mathbf{H}}_{1,k} \\
\vdots \\
\hat{\mathbf{H}}_{t,k}
\end{array}
\right]$. We rewrite Eq.~\eqref{eq6} in trace form as follows:
\begin{align}
\label{eq1:appendix}
\mathop{\arg\!\min}_{W_{\theta_{t,k}^M}}\, \mathrm{tr}&\left[\left(
\bm{\mathcal{H}}_{t,k} - 
\hat{\bm{\mathcal{H}}}_{t,k}W_{\theta_{t,k}^M}
\right)^\top\left(\bm{\mathcal{H}}_{t,k} - 
\hat{\bm{\mathcal{H}}}_{t,k} W_{\theta_{t,k}^M}\right)\right] \nonumber \\ 
&+ \gamma\mathrm{tr}\left(W_{\theta_{t,k}^M}^\top W_{\theta_{t,k}^M}\right).
\end{align}

Further, it can be expanded as follows:
\begin{align}
\label{eq2:appendix}
\mathop{\arg\!\min}_{W_{\theta_{t,k}^M}}\, \mathrm{tr}&\!\left(
\bm{\mathcal{H}}_{t,k}^\top \bm{\mathcal{H}}_{t,k}\!-\!\bm{\mathcal{H}}_{t,k}^\top \hat{\bm{\mathcal{H}}}_{t,k} W_{\theta_{t,k}^M}\!-\!W_{\theta_{t,k}^M}^\top \hat{\bm{\mathcal{H}}}_{t,k}^\top\bm{\mathcal{H}}_{t,k} \right. \nonumber \\ 
&\left. +W_{\theta_{t,k}^M}^\top \hat{\bm{\mathcal{H}}}_{t,k}^\top\hat{\bm{\mathcal{H}}}_{t,k} W_{\theta_{t,k}^M}+\gamma W_{\theta_{t,k}^M}^\top W_{\theta_{t,k}^M}
\right).
\end{align}
Convexity guarantees that the optimal solution is achieved at a stationary point with respect to $W_{\theta_{t,k}^M}$. Accordingly, the gradient of the above equation is derived as:
\begin{align}
&\frac{\partial}{\partial W_{\theta_{t,k}^M}}\mathrm{tr}\left(
\bm{\mathcal{H}}_{t,k}^\top \bm{\mathcal{H}}_{t,k}-\bm{\mathcal{H}}_{t,k}^\top \hat{\bm{\mathcal{H}}}_{t,k} W_{\theta_{t,k}^M}-W_{\theta_{t,k}^M}^\top \hat{\bm{\mathcal{H}}}_{t,k}^\top\bm{\mathcal{H}}_{t,k} \right. \nonumber \\
&\qquad\qquad\left. +W_{\theta_{t,k}^M}^\top \hat{\bm{\mathcal{H}}}_{t,k}^\top\hat{\bm{\mathcal{H}}}_{t,k} W_{\theta_{t,k}^M}+\gamma W_{\theta_{t,k}^M}^\top W_{\theta_{t,k}^M}
\right) \nonumber\\
&= -\frac{\partial \mathrm{tr}\left(\bm{\mathcal{H}}_{t,k}^\top \hat{\bm{\mathcal{H}}}_{t,k} W_{\theta_{t,k}^M}\right)}{\partial W_{\theta_{t,k}^M}} - 
\frac{\partial \mathrm{tr}\left(W_{\theta_{t,k}^M}^\top \hat{\bm{\mathcal{H}}}_{t,k}^\top\bm{\mathcal{H}}_{t,k}\right)}{\partial W_{\theta_{t,k}^M}} \nonumber\\
&\quad\:+ \frac{\partial \mathrm{tr}\left(W_{\theta_{t,k}^M}^\top \hat{\bm{\mathcal{H}}}_{t,k}^\top\hat{\bm{\mathcal{H}}}_{t,k} W_{\theta_{t,k}^M}\right)}{\partial W_{\theta_{t,k}^M}} + 
\gamma\frac{\partial \mathrm{tr}\left(W_{\theta_{t,k}^M}^\top W_{\theta_{t,k}^M}\right)}{\partial W_{\theta_{t,k}^M}} \nonumber\\
&= - \hat{\bm{\mathcal{H}}}_{t,k}^\top \bm{\mathcal{H}}_{t,k} -
\hat{\bm{\mathcal{H}}}_{t,k}^\top\bm{\mathcal{H}}_{t,k} +
\hat{\bm{\mathcal{H}}}_{t,k}^\top\hat{\bm{\mathcal{H}}}_{t,k}W_{\theta_{t,k}^M} \nonumber\\
&\quad\:+ \hat{\bm{\mathcal{H}}}_{t,k}^\top\hat{\bm{\mathcal{H}}}_{t,k}W_{\theta_{t,k}^M} + 2\gamma W_{\theta_{t,k}^M}\nonumber\\
&= -2\hat{\bm{\mathcal{H}}}_{t,k}^\top \bm{\mathcal{H}}_{t,k}+2\hat{\bm{\mathcal{H}}}_{t,k}^\top\hat{\bm{\mathcal{H}}}_{t,k}W_{\theta_{t,k}^M} + 2\gamma W_{\theta_{t,k}^M}. \label{eq3:appendix}
\end{align}
Let Eq.~\eqref{eq3:appendix} equal zero, from which we obtain:
\begin{align}
\label{eq4:appendix}
W_{\theta_{t,k}^M}=&\left(\hat{\bm{\mathcal{H}}}_{t,k}^\top\hat{\bm{\mathcal{H}}}_{t,k}+\gamma \mathbf{I}\right)^{-1}\hat{\bm{\mathcal{H}}}_{t,k}^\top \bm{\mathcal{H}}_{t,k} \nonumber\\
=& \left(\sum_{i=0}^{t}\hat{\mathbf{H}}_{i,k}^\top \hat{\mathbf{H}}_{i,k}+\gamma \mathbf{I}\right)^{-1}\sum_{i=0}^{t}\hat{\mathbf{H}}_{i,k}^\top \mathbf{H}_{i,k}.
\end{align}
By this point, the closed-form solution of Eq.~\eqref{eq6} has been rigorously derived.

Notably, Eq.~\eqref{eq6} can be regarded as a joint learning objective, endowing ADR with absolute resistance to feature drift. HAM inherits NECGL properties and executes without historical examples, ensuring data privacy.

\subsection{Analytic Classifier Reconstruction}

In the Incremental Task Adaptation phase, GNNs are fine-tuned exclusively on data from task $\mathcal{T}_t$, rendering the linear classifier $g_{\phi_t}(\cdot)$ highly biased and incompatible with the merged encoder $\mathcal{F}_{\theta_t^M}(\cdot)$. To this end, we propose ACR to enable the re-learning of the classifier.

A central issue arises from the persistent restriction on historical data access, such that freezing the merged model and fine-tuning the linear classifier perpetuates recency bias~\cite{zhao2020maintaining, song2025exploring}, resulting in severe catastrophic forgetting. We are motivated by the joint ridge regression formulation and establish the following optimization problem:
\begin{equation}
\label{eq8}
\mathop{\arg\!\min}_{W_{\phi_{t}^M}}\left\|\left[
\begin{array}{c}
\mathbf{Y}_{0} \\
\mathbf{Y}_{1} \\
\vdots \\
\mathbf{Y}_{t}
\end{array}
\right] - 
\left[
\begin{array}{c}
\mathbf{H}_{0}^M \\
\mathbf{H}_{1}^M \\
\vdots \\
\mathbf{H}_{t}^M
\end{array}
\right]W_{\phi_{t}^M}
\right\|_{F}^2 + \gamma\left\|W_{\phi_{t}^M}\right\|_F^2,
\end{equation}
where $\mathbf{Y}_t$ is the one-hot label matrix for task $\mathcal{T}_t$, $\mathbf{H}_t^M$ represents the node embedding matrix from the merged encoder $\mathcal{F}_{\theta^M_t}(\cdot)$, and $W_{\phi_{t}^{M}}$ specifies the parameters of the linear classifier $g_{\phi_{t}^{M}}(\cdot)$ to be learned. Let the gradient of Eq.~\eqref{eq8} with respect to $W_{\phi_{t}^{M}}$ be zero. Which leads to:
\begin{equation}
\label{eq9}
W_{\phi_{t}^M}^*=\left(\sum_{i=0}^{t}{\mathbf{H}_{i}^M}^\top \mathbf{H}_{i}^M+\gamma \mathbf{I}\right)^{-1}\sum_{i=0}^{t}{\mathbf{H}_{i}^M}^\top \mathbf{Y}_{i}.
\end{equation}
As with the encoder-side memory bank, keeping only $R_{t}^\phi\!=\!\sum_{i=0}^{t}{\mathbf{H}_i^M}^\top\mathbf{H}_{i}^{M}$ and $Q_t^\phi\!=\!\sum_{i=0}^{t}{\mathbf{H}_i^M}^\top\mathbf{Y}_i$ enables incremental classifier updates without iterative BP, incurring a total memory cost of $\mathcal{O}\left({d_{K-1}^{out}}^2+d_{K-1}^{out}\sum_{i=0}^{t}c_i\right)$. Moreover, following Cover's theorem~\cite{cover2006geometrical}, we execute an $\alpha$-fold feature expansion on $\{\mathbf{H}_{i}^M\}_{i=0}^{t}$ to improve the linear separability.
\begin{equation}
\label{eq10}
\mathbf{H}_{i}^{\mathcal{B}}=\mathcal{B}(\mathbf{H}_{i}^M)=\sigma(\mathbf{H}_{i}^M W_\psi),
\end{equation}
where $\mathcal{B}(\cdot)$ denotes the feature buffer layer, comprising a randomly parameterized feature expansion matrix $W_\psi\!\in\!\mathbb{R}^{d_{K-1}^{out}\times (\alpha\times d_{K-1}^{out})}$ and a nonlinear activation function $\sigma(\cdot)$. Substituting $\{\mathbf{H}_i^M\}_{i=0}^{t}$ in Eq.~\eqref{eq8} with $\{\mathbf{H}_i^\mathcal{B}\}_{i=0}^{t}$, Eq.~\eqref{eq9} can be reformulated as follows:
\begin{equation}
\label{eq11}
W_{\phi_{t}^M}^*=\left(\sum_{i=0}^{t}{\mathbf{H}_{i}^\mathcal{B}}^\top \mathbf{H}_{i}^\mathcal{B}+\gamma \mathbf{I}\right)^{-1}\sum_{i=0}^{t}{\mathbf{H}_{i}^\mathcal{B}}^\top \mathbf{Y}_{i}.
\end{equation}
The memory footprint of the classifier-side memory bank expands to $\mathcal{O}\left(\alpha^2 {d_{K-1}^{out}}^2+\alpha d_{K-1}^{out}\sum_{i=0}^{t}c_i\right)$. Eqs.~\eqref{eq9} and \eqref{eq11} follow the derivation of Eq.~\eqref{eq7}.

Analogous to HAM, ACR provides privacy guarantees and is formally equivalent to joint learning, theoretically establishing a zero-forgetting paradigm. We present the training pseudocode for the proposed ADR in Algorithm~\ref{alg}.

\begin{algorithm}[t]
  \caption{Training Pipeline of Our ADR}
  \label{alg}
  \begin{algorithmic}
    \STATE {\bfseries Input:} Task stream $\mathcal{T}=\{\mathcal{T}_0, \mathcal{T}_1,\dots,\mathcal{T}_{N-1}\}$, graph encoder $\mathcal{F}_{\theta_t}(\cdot)$, linear classifier $g_{\phi_t}(\cdot)$, feature buffer layer $\mathcal{B}(\cdot)$, encoder-side memory bank $\mathcal{M}_{\mathcal{F}}$, classifier-side memory bank $\mathcal{M}_{g}$, learning rate $\eta_t$, maximum epoch $E$.
    \STATE {\bfseries Initialize:} $\mathcal{M}_{\mathcal{F}}, \mathcal{M}_{g}\leftarrow \{\}, \{\}$
    \FOR{task $t=0$ {\bfseries to} $N-1$}
        \STATE \emph{\textcolor{blue}{$\triangleright$ Incremental Task Adaptation}}
        \FOR{epoch $e=1$ {\bfseries to} $E$}
        \STATE Compute $\mathcal{L}_{ce}$ according to Eq.~\eqref{eq2}.
        \STATE $\Theta_t\leftarrow \Theta_t-\eta_t \nabla_{\Theta_t}\mathcal{L}_{ce}$.
    \ENDFOR    
    \STATE \emph{\textcolor{blue}{$\triangleright$ Hierarchical Analytic Merging}}
    \STATE Merge layer-wise linear transformations in $\mathcal{F}_{\theta_t}(\cdot)$ according to Eqs.~\eqref{eq6} and \eqref{eq7}, obtaining the merged encoder $\mathcal{F}_{\theta_t^M}(\cdot)$.
    \STATE $\mathcal{M}_\mathcal{F}\leftarrow\{(R_{t,k}^\theta, Q_{t,k}^\theta)\}_{k=0}^{K-1}$.
    \STATE \emph{\textcolor{blue}{$\triangleright$ Analytic Classifier Reconstruction}}
    \STATE Feature expansion according to Eq.~\eqref{eq10}.
    \STATE Recompute classifier parameters according to Eqs.~\eqref{eq8} and \eqref{eq11}, obtaining the reconstructed linear classifier $g_{\phi_t^M}(\cdot)$.
    \STATE $\mathcal{M}_g\leftarrow \{R_{t}^\phi,Q_t^\phi\}$.
    \ENDFOR
    \STATE \textbf{return} $\mathcal{F}_{\theta_{N-1}^M}(\cdot)$, $g_{\phi_{N-1}^M}(\cdot)$
  \end{algorithmic}
\end{algorithm}

\section{Experiments}

\subsection{Experimental Setup}

\textbf{Datasets.} We conduct a comprehensive evaluation of the proposed ADR on four established node classification benchmarks: CS-CL~\cite{shchur2018pitfalls}, CoraFull-CL~\cite{mccallum2000automating}, Arxiv-CL~\cite{hu2020open}, and Reddit-CL~\cite{hamilton2017inductive}. These datasets, spanning diverse scales, are systematically arranged into a chronologically ordered task stream $\mathcal{T}=\{\mathcal{T}_0,\mathcal{T}_1,\dots,\mathcal{T}_{N-1}\}$ following~\cite{zhang2022cglb, zhuang2022acil}, with $\mathcal{T}_0$ as the base task, containing nearly half of the data, and $\mathcal{T}_{1:N-1}$ as incremental tasks, each comprising two disjoint classes to emulate the continuous arrival of new tasks. Class-level data are split into training, validation, and test sets in a 6:2:2 ratio. Table~\ref{tab:dataset} summarizes the key statistical characteristics of the four datasets.

\textbf{Implementation Details.} ADR employs a 2-layer GCN~\cite{kipf2016semi} backbone with 128 hidden dimensions and a 0.5 dropout rate. In the Incremental Task Adaptation phase, the model is optimized using Adam, with the base task learning rate set to $\eta_0=1\times 10^{-3}$ and incremental tasks to $\eta_{t>0}=1\times 10^{-4}$, and trained for up to 200 epochs to ensure stable convergence. In solving ridge regression, the regularization weight $\gamma$ and feature expansion factor $\alpha$ are determined via grid search over $\{10^{i}\mid i\in\mathbb{Z}, -3\leq i\leq 0\}$ and $\{2^i\mid i\in\mathbb{Z}, 0\leq i\leq 6\}$, respectively. All experiments run on a single NVIDIA 4090D GPU, with the code implemented based on Continual Graph Learning Benchmark (CGLB)~\cite{zhang2022cglb}. We perform mini-batch training across all datasets with a batch size of 2000. The mean and standard deviation over 5 repeated runs are reported.

\textbf{Global Testing Protocol.} In class-incremental learning, task-specific graphs are typically assumed to be topologically isolated from one another. Accordingly, each of the previous task graphs is independently fed into the trained GNNs for performance assessment, a protocol termed local testing. A recent study~\cite{cheng2025can} has revealed that, in semi-supervised node classification, the co-presence of training and test nodes within the same subgraph allows the task ID to be accurately inferred via na\"{i}ve task prototype matching. This risks collapsing class-incremental learning into task-incremental learning, inflating reported performance. In this paper, we resort to a more rigorous global testing protocol rather than local testing to evaluate the genuine continual learning capability of the proposed method. To be specific, the test graph preserves inter-task edges, with all task-specific graphs consolidated into a single graph for evaluation, thus preventing the task-ID shortcut. Note that changes are applied only to the validation and test phases, while leaving the training pipeline unaffected.

\textbf{Baselines.} We compare ADR with four categories of SOTA methods: regularization-based methods (i.e., LWF~\cite{li2017learning}, EWC~\cite{kirkpatrick2017overcoming}, MAS~\cite{aljundi2018memory}, and TWP~\cite{liu2021overcoming}), rehearsal-based methods (i.e., CaT~\cite{liu2023cat}, and ER-GNN~\cite{zhou2021overcoming}), Non-Exemplar methods (i.e., POLO~\cite{wang2023non}, and EFC~\cite{magistri2024elastic}), and ACL methods (i.e., DPCR~\cite{he2025semantic}, ACIL~\cite{zhuang2022acil}, and DS-AL~\cite{zhuang2024ds}). To ensure fairness, the feature expansion factor $\alpha$ in ACIL and DS-AL is likewise searched within the parameter ranges described above. Moreover, Bare (direct fine-tuning) and Joint (joint training) are implemented to delineate the lower and upper performance bounds.

\begin{table}[t]
\caption{Key statistical characteristics of four node classification benchmarks.}
\label{tab:dataset}
\begin{center}
\begin{sc}
\resizebox{\columnwidth}{!}{
\begin{tabular}{c|cccc}
\toprule
Datasets         & CS-CL  & CoraFull-CL & Arxiv-CL & Reddit-CL \\ \midrule
\# nodes         & 18333  & 19793       & 169343   & 232965    \\
\# edges         & 163788 & 126842      & 1166243  & 114615892 \\
\# features      & 6805   & 8710        & 128      & 602       \\
\# all classes       & 15     & 70          & 40       & 40        \\
\# base classes  & 5      & 30          & 20       & 20        \\
\# incremental classes & 10     & 40          & 20       & 20        \\
\# split         & 5$+$5$\times$2  & 30$+$20$\times$2     & 20$+$10$\times$2  & 20$+$10$\times$2   \\
\# tasks         & 6      & 21          & 11       & 11       \\ \bottomrule
\end{tabular}
}
      \end{sc}
  \end{center}
\end{table}

\textbf{Evaluation Metrics.} Following \cite{zhuang2024ds, he2025semantic}, we quantify the continual learning performance of all methods via average incremental accuracy $\mathcal{A}_{avg}$ and final accuracy $\mathcal{A}_f$, which are defined as follows:
\begin{equation}
    \label{eq12}
    \mathcal{A}_{avg}=\frac{1}{N}\sum_{t=0}^{N-1}\mathcal{A}_t, \quad\mathcal{A}_f=\frac{\sum_{t=0}^{N-1}\mathbf{M}_{N-1,t}}{N},
\end{equation}
where $\mathcal{A}_t=\frac{\sum_{i=0}^{t}\mathbf{M}_{t,i}}{t+1}$, with $\mathbf{M}$ denoting a lower-triangular performance matrix, and $\mathbf{M}_{t,i}$ reflecting the test accuracy on task $\mathcal{T}_i$ after training on task $\mathcal{T}_t$. Moreover, we quantify model plasticity via learning accuracy $\mathcal{A}_l$~\cite{riemer2018learning}, defined as follows:
\begin{equation}
    \label{eq13}
    \mathcal{A}_l=\frac{1}{N}\sum_{t=0}^{N-1}\mathbf{M}_{t,t}.
\vspace{-12pt}
\end{equation}

\begin{table*}[t]
  \caption{Performance comparison on the four benchmarks. The best results are highlighted in bold, and the second-best results are underlined.}
  \label{tab1}
  \begin{center}
      \begin{sc}
      \resizebox{\textwidth}{!}{
        \begin{tabular}{c|cc|cc|cc|cc}
        \toprule
\multirow{2}{*}{Methods} & \multicolumn{2}{c|}{CS-CL}        & \multicolumn{2}{c|}{CoraFull-CL} & \multicolumn{2}{c|}{Arxiv-CL}  & \multicolumn{2}{c}{Reddit-CL} \\ \cline{2-9}
                                  & $\mathcal{A}_{avg}$(\%)      & $\mathcal{A}_f$(\%)      & $\mathcal{A}_{avg}$(\%)      & $\mathcal{A}_f$(\%)     & $\mathcal{A}_{avg}$(\%)      & $\mathcal{A}_f$(\%)   & $\mathcal{A}_{avg}$(\%)      & $\mathcal{A}_f$(\%)   \\ \midrule
Joint                    & -                   & 88.92$\pm$0.25          & -                   & 61.62$\pm$0.26         & -                   & 42.11$\pm$0.20       & -                   & 79.82$\pm$0.69       \\
Bare                     & 53.24$\pm$1.83          & 34.54$\pm$6.39          & 17.55$\pm$0.15          & 3.75$\pm$0.90          & 18.16$\pm$0.10          & 8.99$\pm$0.02        & 40.47$\pm$0.91          & 12.64$\pm$1.97       \\ \midrule
LWF                      & 62.59$\pm$1.40          & 38.40$\pm$2.30          & 18.53$\pm$0.75          & 6.91$\pm$1.17          & 17.64$\pm$0.18          & 8.97$\pm$0.05        & 37.73$\pm$0.76          & 12.76$\pm$2.45       \\
EWC                      & 72.76$\pm$3.04          & 49.67$\pm$5.28          & 20.23$\pm$0.40          & 5.44$\pm$1.92          & 17.01$\pm$0.15          & 8.91$\pm$0.03        & 40.93$\pm$0.74          & 8.82$\pm$0.04        \\
MAS                      & 75.22$\pm$2.02          & 66.91$\pm$2.09          & 19.21$\pm$0.29          & 9.09$\pm$1.83          & 17.68$\pm$0.49          & 8.81$\pm$0.51        & 53.02$\pm$3.55          & 33.11$\pm$4.24       \\
TWP                      & 55.89$\pm$0.53          & 41.95$\pm$3.18          & 19.78$\pm$1.12          & 8.67$\pm$2.23          & 17.03$\pm$0.20          & 8.95$\pm$0.16        & 39.79$\pm$0.92          & 9.97$\pm$0.41        \\
CaT                      & 63.85$\pm$1.52          & 51.16$\pm$3.09          & 27.38$\pm$1.29          & 11.02$\pm$1.73         & 15.42$\pm$0.25          & 8.90$\pm$0.03        & 51.95$\pm$1.80          & 22.85$\pm$1.47       \\
ER-GNN                   & 84.05$\pm$0.42          & 80.87$\pm$0.52          & 20.55$\pm$0.26          & 4.91$\pm$0.97          & 29.47$\pm$0.33          & 16.76$\pm$0.24       & 81.66$\pm$0.44 & 69.70$\pm$1.43       \\
POLO                     & 68.60$\pm$1.47          & 41.62$\pm$3.35          & 26.63$\pm$1.53          & 11.10$\pm$1.80         & 14.76$\pm$0.10          & 4.56$\pm$0.02        & 55.03$\pm$2.45          & 24.77$\pm$1.02       \\
EFC                      & 72.28$\pm$1.23          & 61.25$\pm$2.97          & 23.01$\pm$0.40          & 7.64$\pm$0.58          & 24.10$\pm$0.31          & 12.81$\pm$0.42       & 78.45$\pm$0.71          & 69.91$\pm$0.75       \\
DPCR                     & 63.78$\pm$1.26          & 43.89$\pm$3.13          & 32.86$\pm$0.88          & 7.12$\pm$1.08          & 31.30$\pm$0.84          & 23.67$\pm$1.07       & 66.09$\pm$2.86          & 38.27$\pm$2.50       \\
ACIL                     & 88.58$\pm$0.25          & \underline{85.57$\pm$0.59}          & 61.49$\pm$0.19          & 51.92$\pm$0.58         & 39.37$\pm$0.20          & \underline{35.08$\pm$0.23}       & 81.37$\pm$0.39          & 73.93$\pm$0.53       \\
DS-AL                    & \underline{88.88$\pm$0.42}    & 85.54$\pm$0.96    & \underline{61.78$\pm$0.59}    & \underline{52.69$\pm$0.65}   & \textbf{41.41$\pm$0.24} & \textbf{36.78$\pm$0.30} & \underline{81.77$\pm$0.25}          & \underline{74.40$\pm$0.33} \\
\rowcolor{blue!8} ADR               & \textbf{90.71$\pm$0.22} & \textbf{87.21$\pm$0.42} & \textbf{67.52$\pm$0.83}                    & \textbf{60.03$\pm$1.22}                   & \underline{39.66$\pm$0.43}                    & 34.74$\pm$0.21                 & \textbf{82.52$\pm$0.21}                    & \textbf{75.45$\pm$0.40}                \\ \bottomrule
\end{tabular}
}
      \end{sc}
  \end{center}
\end{table*}

\begin{table*}[t]
  \caption{Ablation studies of HAM on the four benchmarks. The best results are highlighted in bold.}
  \label{tab:ablation}
  \begin{center}
      \begin{sc}
      \resizebox{\textwidth}{!}{
\begin{tabular}{c|cc|cc|cc|cc}
\toprule
\multirow{2}{*}{Variants} & \multicolumn{2}{c|}{CS-CL} & \multicolumn{2}{c|}{CoraFull-CL} & \multicolumn{2}{c|}{Arxiv-CL} & \multicolumn{2}{c}{Reddit-CL} \\ \cline{2-9}
                          & $\mathcal{A}_{avg}$(\%)       & $\mathcal{A}_{f}$(\%)       & $\mathcal{A}_{avg}$(\%)          & $\mathcal{A}_{f}$(\%)          & $\mathcal{A}_{avg}$(\%)         & $\mathcal{A}_{f}$(\%)       & $\mathcal{A}_{avg}$(\%)         & $\mathcal{A}_{f}$(\%)         \\ \midrule
w/o HAM                   & 52.28$\pm$1.27  & 36.63$\pm$2.92  & 17.36$\pm$0.19     & 7.65$\pm$0.99      & 20.07$\pm$0.06    & 9.01$\pm$0.02    & 43.83$\pm$0.67    & 17.43$\pm$1.75    \\
w/ Simple Averaging       & 65.08$\pm$1.31  & 48.79$\pm$1.91  & 31.83$\pm$0.48     & 24.75$\pm$0.95     & 23.23$\pm$0.20    & 15.17$\pm$0.37   & 69.29$\pm$2.12    & 48.23$\pm$1.89    \\
w/ Fisher Merging         & 69.07$\pm$2.58  & 58.01$\pm$5.32  & 27.68$\pm$0.66     & 18.64$\pm$1.39     & 26.06$\pm$1.33    & 10.15$\pm$0.23   & 68.53$\pm$2.78    & 48.42$\pm$3.25    \\
w/ MAGMAX Merging         & 67.23$\pm$0.78  & 57.34$\pm$3.98  & 31.08$\pm$1.16     & 21.91$\pm$1.82     & 21.66$\pm$0.24    & 9.64$\pm$0.20    & 70.15$\pm$1.03    & 50.81$\pm$3.52    \\
\rowcolor{blue!8} ADR                       & \textbf{90.71$\pm$0.22}  & \textbf{87.21$\pm$0.42}  & \textbf{67.52$\pm$0.83}     & \textbf{60.03$\pm$1.22}     & \textbf{39.66$\pm$0.43}    & \textbf{34.74$\pm$0.21}   & \textbf{82.52$\pm$0.21}    & \textbf{75.45$\pm$0.40}   \\ \bottomrule
\end{tabular}
}
      \end{sc}
  \end{center}
\end{table*}

\begin{figure}[t]
  \begin{center}
    \centerline{\includegraphics[width=0.9\columnwidth, trim=24 24 21 21, clip]{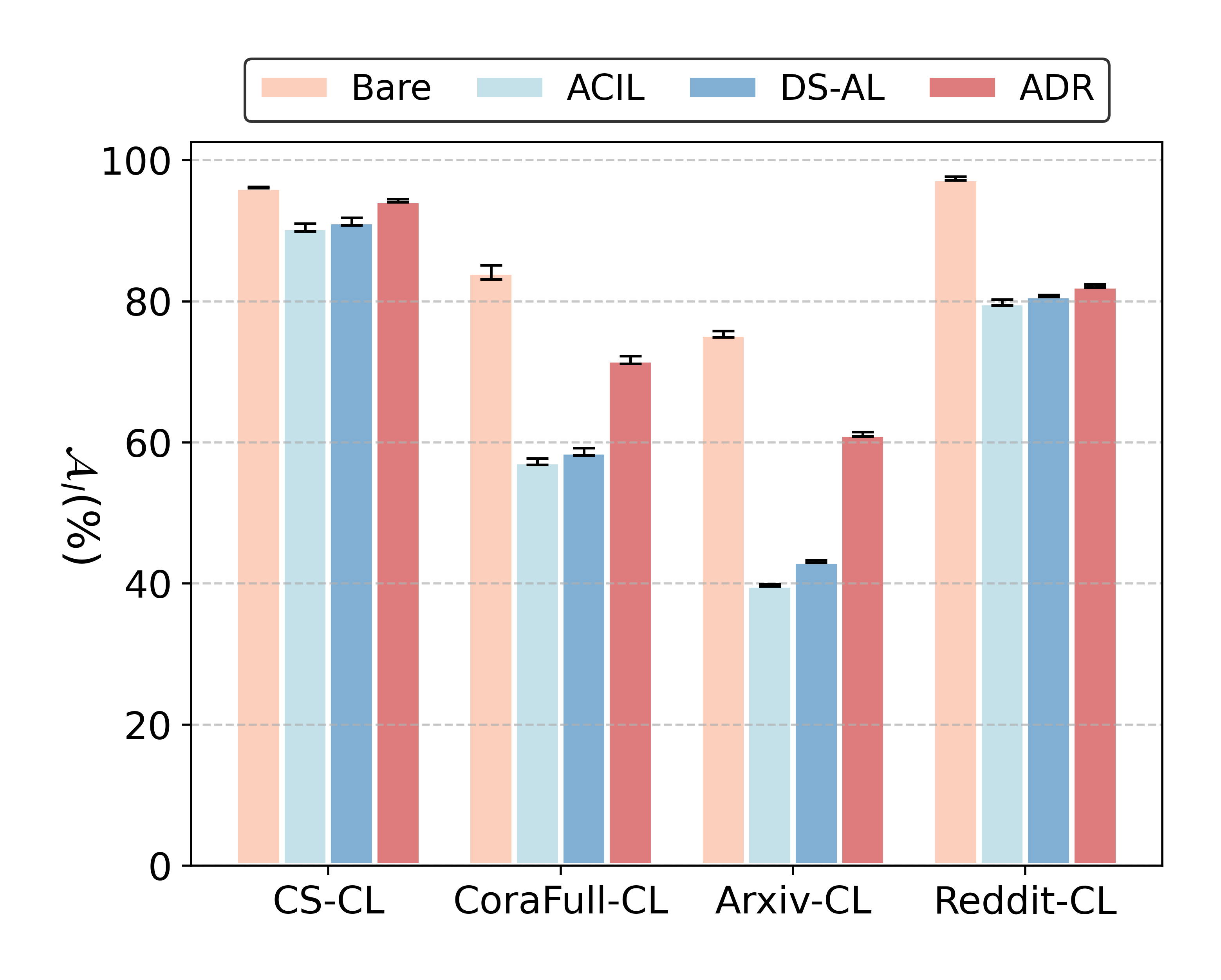}}
    \caption{
    Model plasticity comparison of our ADR versus existing ACL methods on the four benchmarks.
    }
    \label{fig:la}
  \end{center}
\end{figure}

\subsection{Comparison with State-of-the-Art}
\label{sub:Comparison}

\textbf{Overall Performance Analysis.} Table~\ref{tab1} reports a comparison between the proposed ADR and SOTA methods from four categories. Regularization-based methods exhibit inferior performance, as the imposed regularization, while retaining historical knowledge, inevitably constrains the flexibility of the model in adapting to new tasks. Rehearsal-based methods periodically revisit historical examples to reactivate past memory, outperforming regularization-based methods while raising data privacy concerns. CaT alleviates this issue via dataset condensation, yet its efficacy is inherently dependent on the fidelity of the synthetic replay graphs. Non-Exemplar methods cache gaussian prototypes in lieu of raw exemplars, circumventing infringements on privacy, yet feature drift induces notable representational errors. ACL methods achieve superior performance, as recursive LS solutions are theoretically equivalent to joint training. However, such methods typically freeze the pre-trained encoder to guarantee the validity of the linear regression solution, severely compromising model plasticity. DPCR addresses this issue by unfreezing the model on new tasks, yet the ensuing feature drift further degrades performance. Although post-hoc drift compensation mitigates this effect, estimating the shift of prior prototypes from new task graphs is inherently biased, particularly when task distributions differ significantly. Thus, DPCR underperforms ACIL and DS-AL. The proposed ADR capitalizes on HAM to achieve absolute resistance to feature drift while improving model plasticity, delivering competitive performance across all four benchmarks. Moreover, we observe that ADR records marginally lower $\mathcal{A}_f$ than ACIL and DS-AL on Arxiv-CL. This discrepancy arises predominantly from severe class imbalance in certain tasks, which lies beyond the focus of this paper. A detailed analysis is provided in Section~\ref{appendix:d}.

\textbf{Model Plasticity Analysis.} We evaluate the model plasticity of our ADR and existing ACL methods, as shown in Fig.~\ref{fig:la}. Bare, the upper bound of plasticity, is also considered. ACIL and DS-AL demonstrate significantly lower plasticity than Bare, particularly on CoraFull-CL and Arxiv-CL, owing to encoder freezing. The proposed ADR relaxes the frozen parameter constraint, allowing GNNs to adapt flexibly to new task graphs via iterative BP, markedly improving model plasticity.

\subsection{Ablation Studies}

To validate the effectiveness of HAM, a pivotal component of ADR, we conduct ablation studies presented in Table~\ref{tab:ablation}, with the findings summarized as follows: (\romannumeral1) Removing HAM leads to severe feature drift in the classifier-side memory bank $\{R_{t-1}^\phi, Q_{t-1}^\phi\}$, resulting in the linear classifier weights being derived from an invalid ridge regression objective. (\romannumeral2) Replacing HAM with three SOTA model merging methods---Simple Averaging~\cite{choshen2022fusing, wortsman2022model}, Fisher Merging~\cite{matena2022merging}, and MAGMAX Merging~\cite{marczak2024magmax}---induces a substantial performance deterioration relative to ADR, owing to the theoretical equivalence of HAM to the joint learning paradigm, which confers absolute resistance to feature drift.

\begin{figure*}[t]
    \subfloat{
        \includegraphics[width=0.32\textwidth, viewport=165 145 2950 2440, clip]{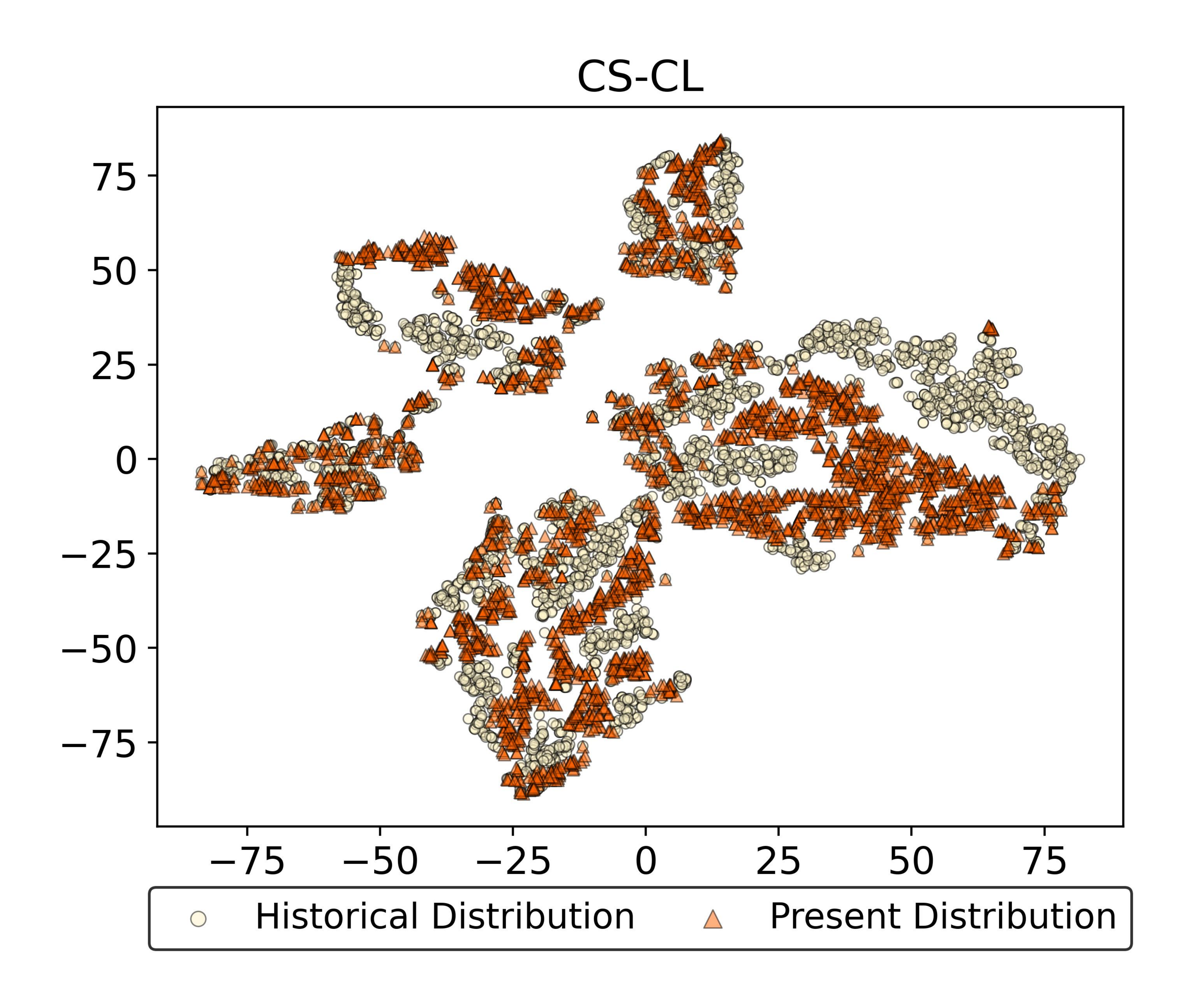}
        \label{fig:fd1}
    }
    \subfloat{
        \includegraphics[width=0.32\textwidth, viewport=165 145 2950 2440, clip]{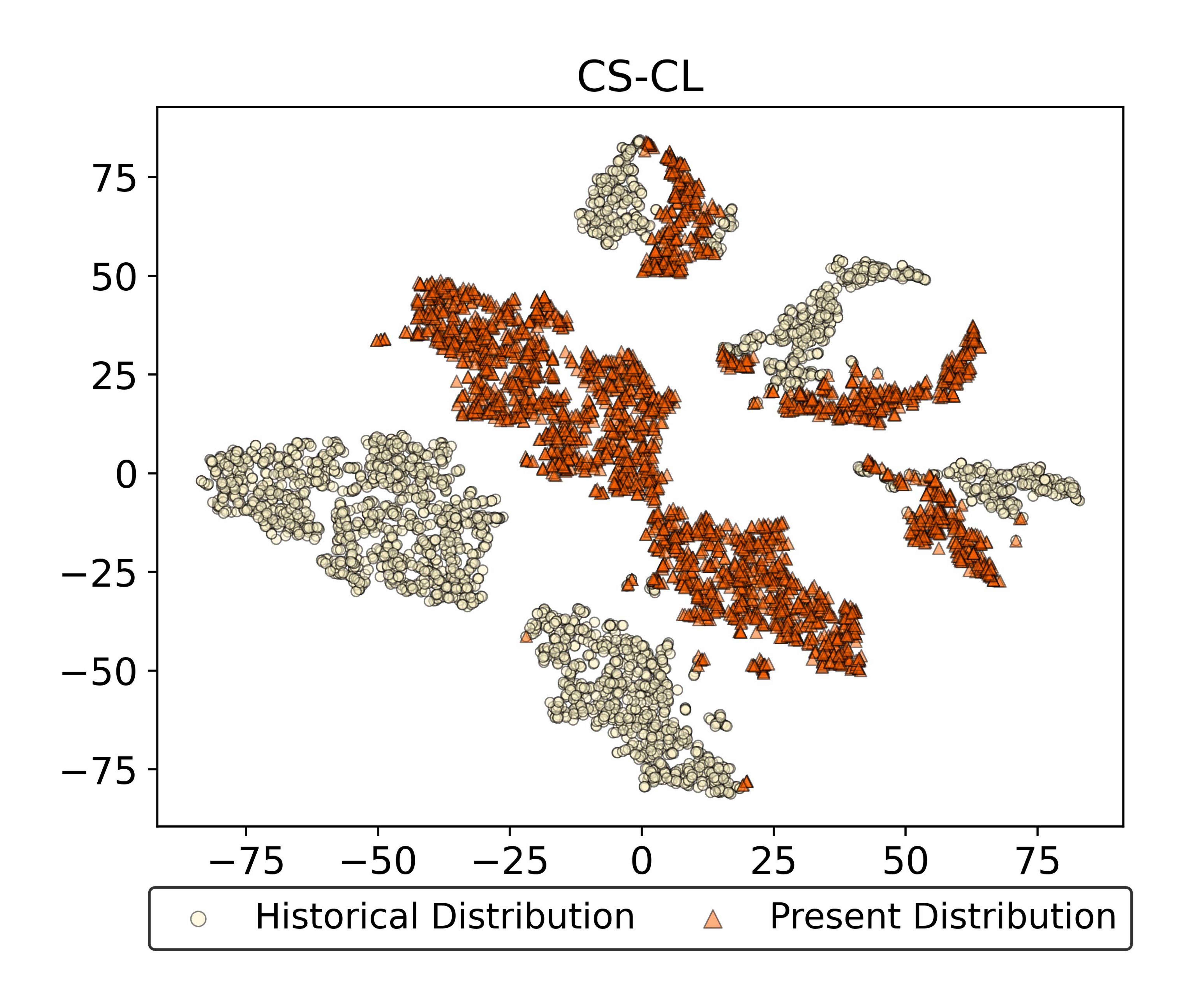}
        \label{fig:fd2}
    }
    \subfloat{
        \includegraphics[width=0.32\textwidth, viewport=165 145 2950 2440, clip]{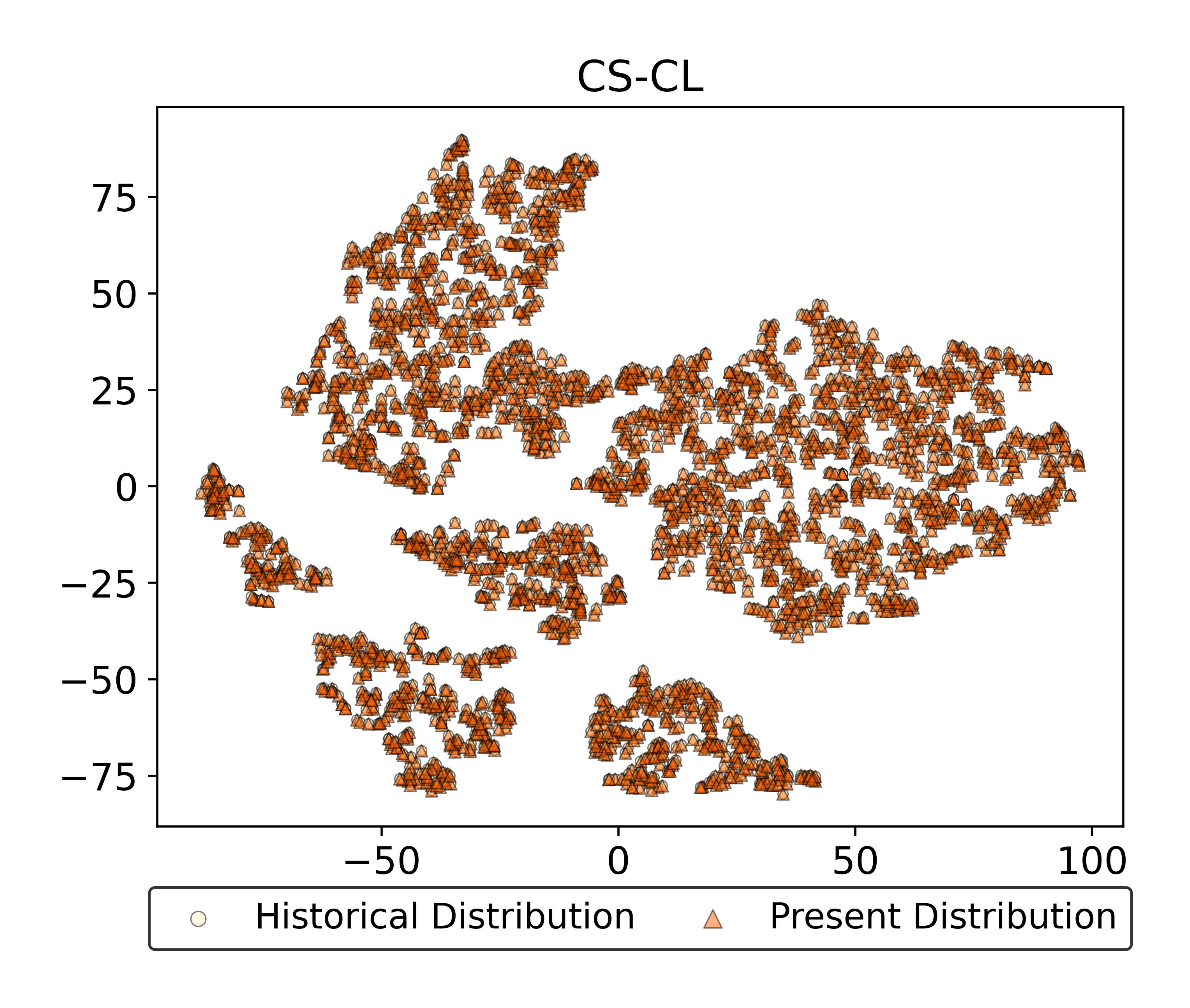}
        \label{fig:fd3}
    }

    \setcounter{subfigure}{0} 
    \captionsetup[subfloat]{labelformat=parens}
    
    \subfloat[EFC]{
        \includegraphics[width=0.32\textwidth, viewport=165 145 2950 2440, clip]{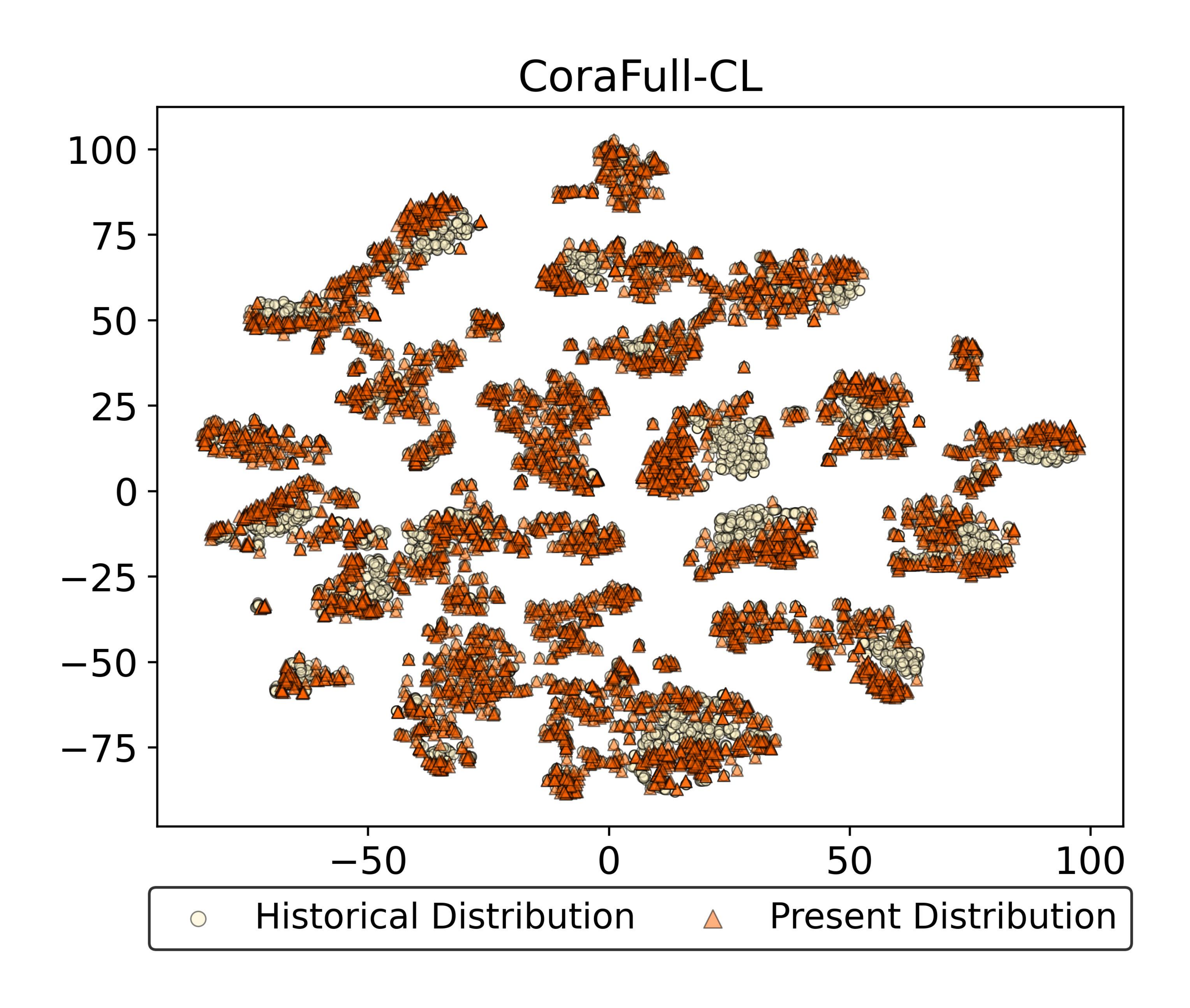}
        \label{fig:fd4}
    }
    \subfloat[DPCR]{
        \includegraphics[width=0.32\textwidth, viewport=165 145 2950 2440, clip]{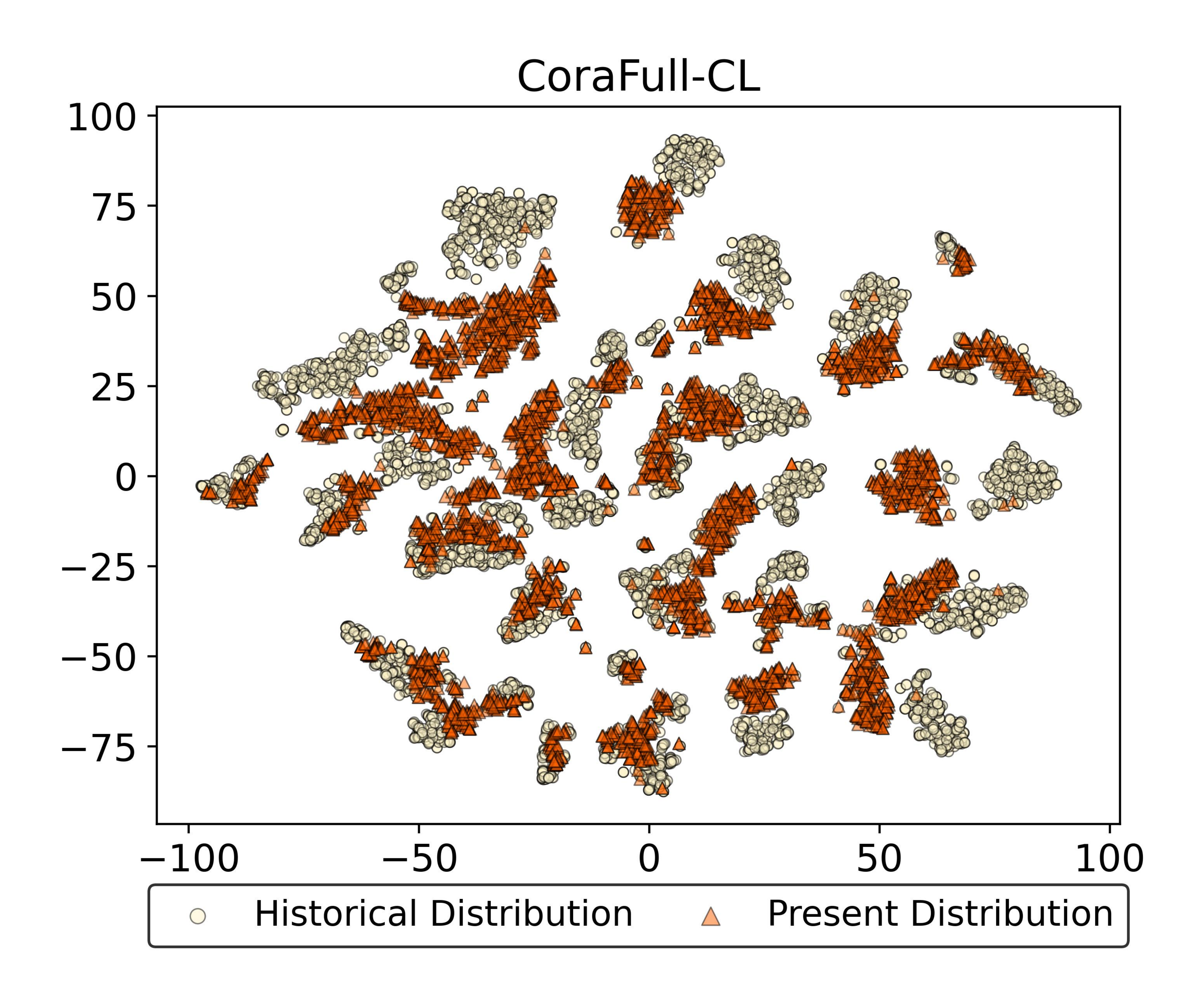}
        \label{fig:fd5}
    }
    \subfloat[ADR]{
        \includegraphics[width=0.32\textwidth, viewport=155 145 3000 2440, clip]{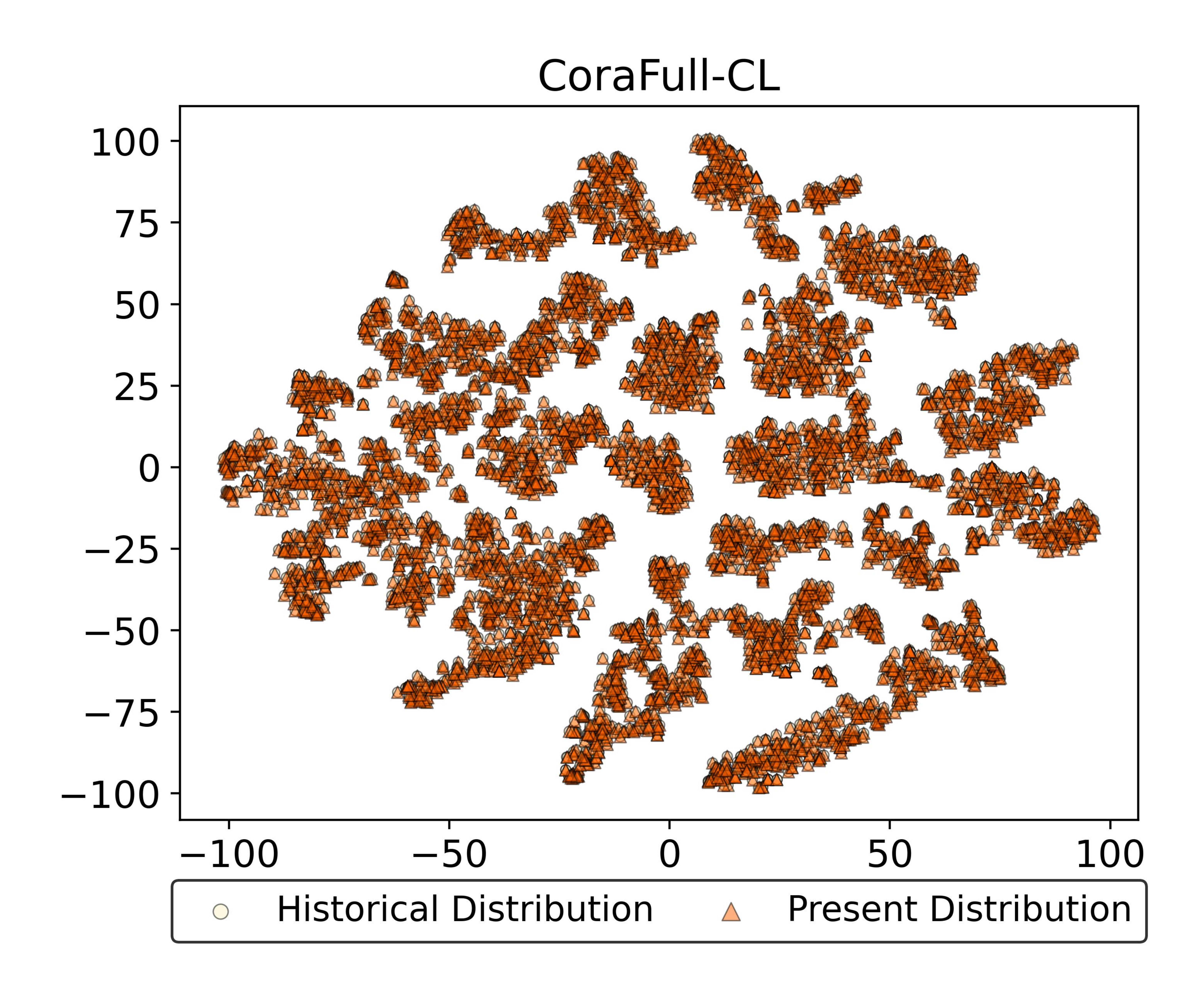}
        \label{fig:fd6}
    }
    
    \caption{Feature drift visualization of EFC, DPCR, and ADR on the base task graph $\mathcal{G}_0$ across CS-CL and CoraFull-CL benchmarks.}
    \label{fig:drift}
\end{figure*}

To intuitively illustrate the absolute resistance of ADR to feature drift, we visualize the distributions of node embeddings for the base task graph $\mathcal{G}_0$, generated by encoders $\mathcal{F}_{\theta_{0}^M}(\cdot)$ and $\mathcal{F}_{\theta_{N-1}^M}(\cdot)$, on the CS-CL and CoraFull-CL benchmarks. In Fig.~\ref{fig:drift}, light-hued circles and dark-hued triangles indicate the outputs of the two encoders, respectively. It can be observed that the Non-Exemplar method EFC and the unfrozen ACL method DPCR exhibit severe feature shift after learning multiple tasks, whereas ADR remains fully immune.

\begin{figure*}[t]
  \begin{center}
    \subfloat{
        \includegraphics[width=0.232\textwidth, trim=0 23 21 22, clip]{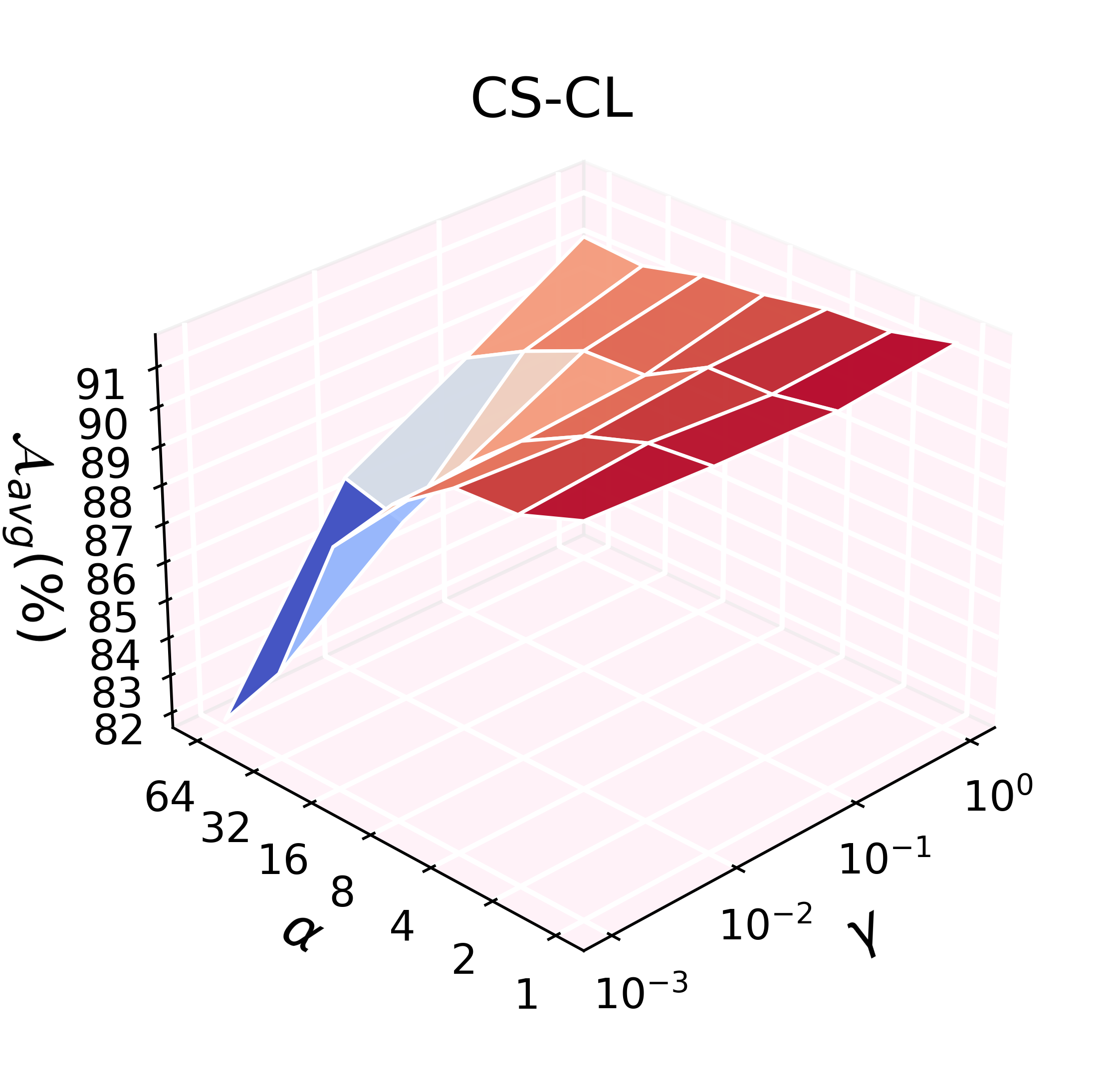}
        \label{fig:param1}
    }
    \hfil
    \subfloat{
        \includegraphics[width=0.232\textwidth, trim=0 23 21 22, clip]{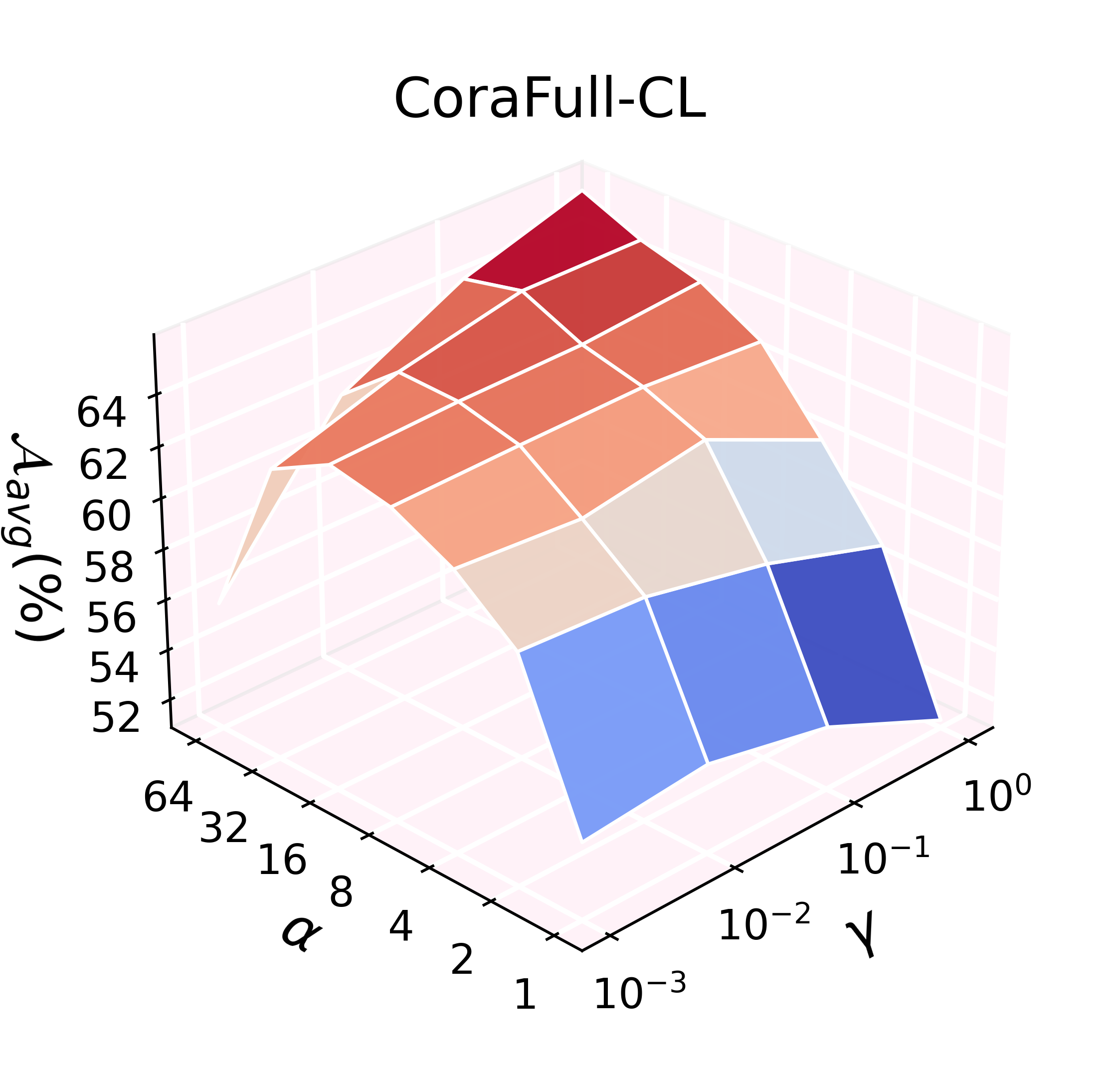}
        \label{fig:param2}
    }
    \hfil
    \subfloat{
        \includegraphics[width=0.232\textwidth, trim=0 23 21 22, clip]{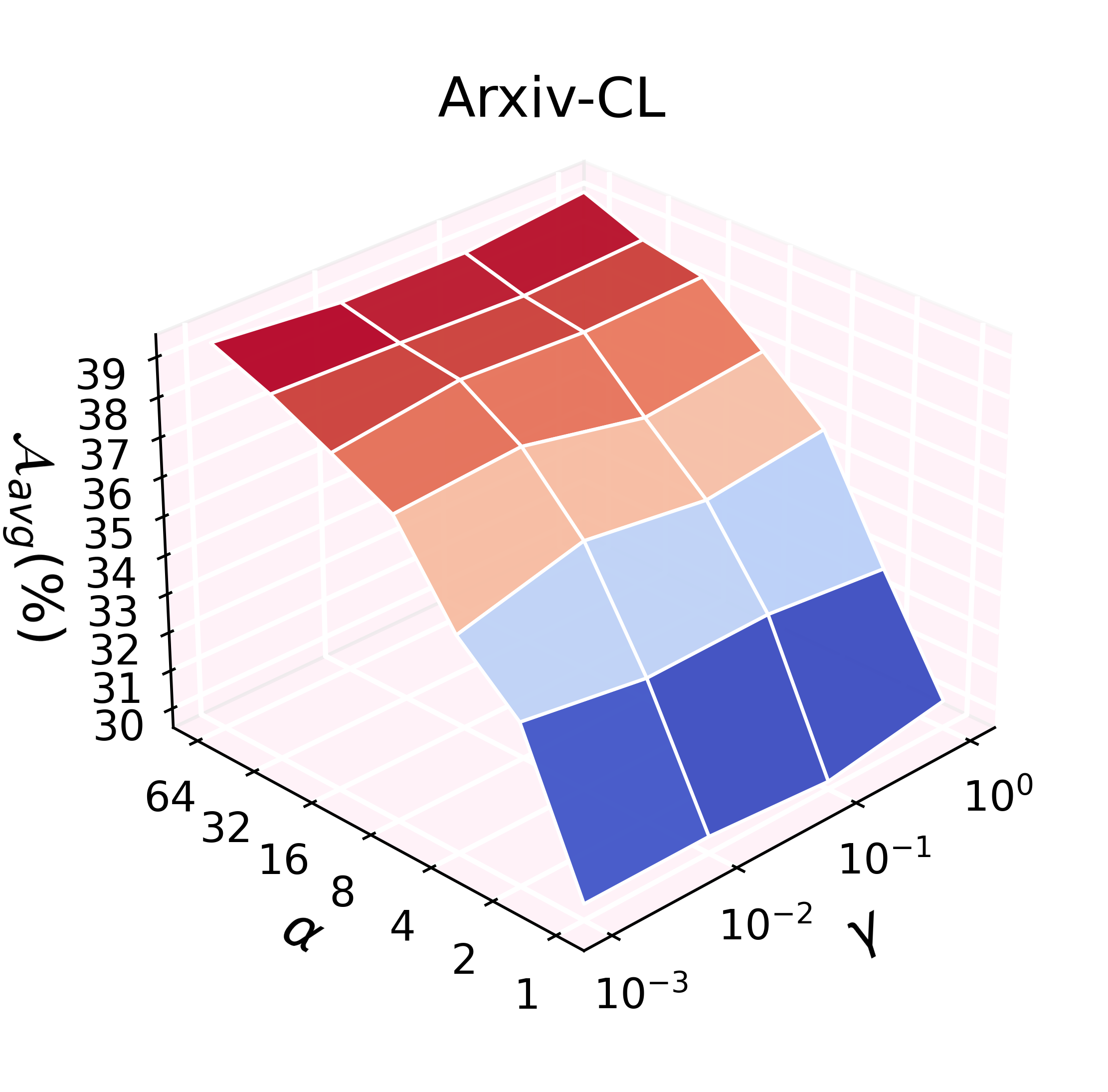}
        \label{fig:param3}
    }
    \hfil
    \subfloat{
        \includegraphics[width=0.232\textwidth, trim=0 23 21 22, clip]{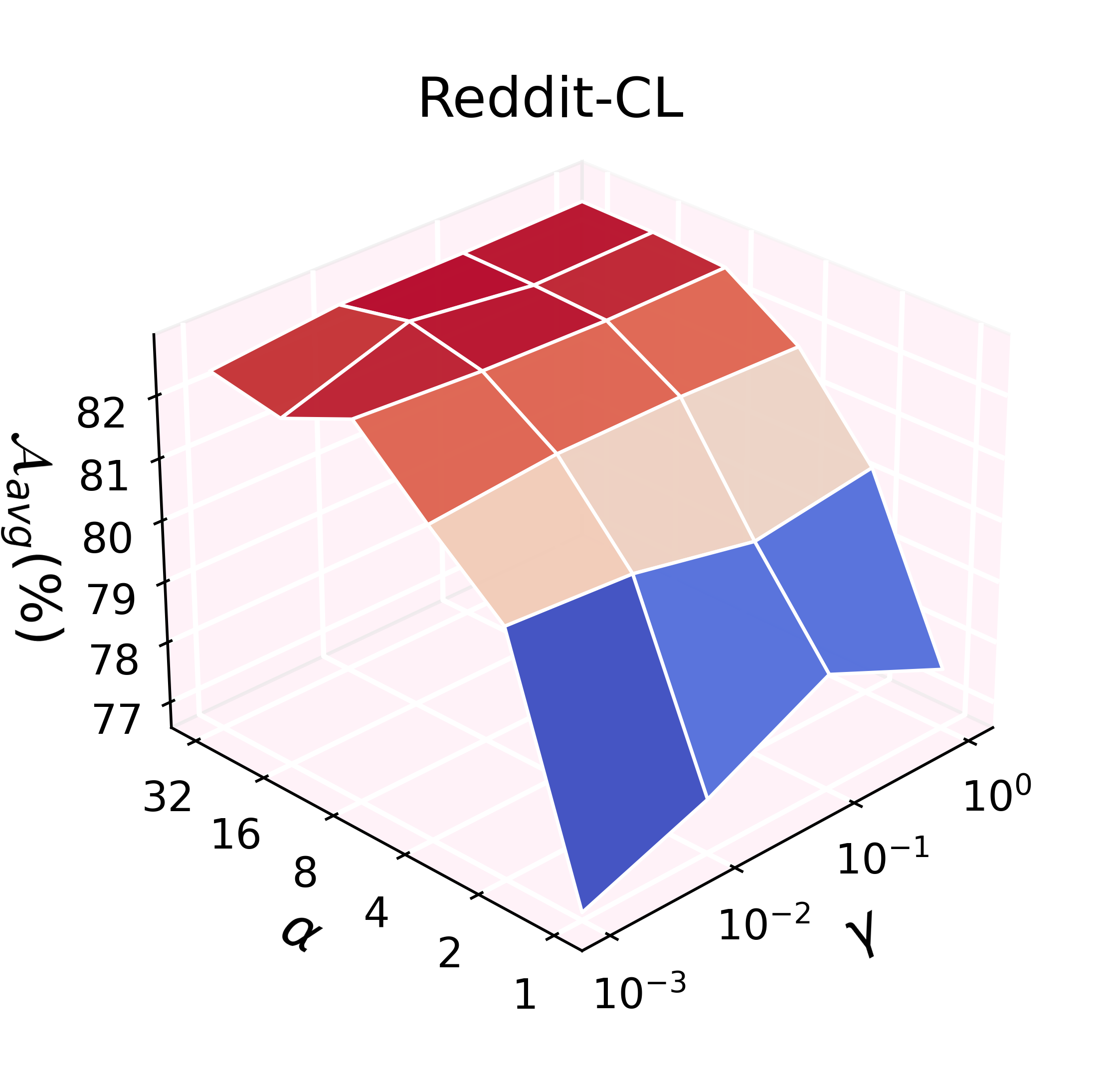}
        \label{fig:param4}
    }
    \caption{Visualization of the grid search over hyperparameters $\alpha$ and $\gamma$, with $\mathcal{A}_{avg}$ reported.}
    \label{fig:params}
  \end{center}
\end{figure*}

\subsection{Parameter Analysis}

A grid search over the feature expansion factor $\alpha\!\in\!\{2^i\mid i\in\mathbb{Z}, 0\leq i\leq 6\}$ and regularization weight $\gamma\!\in\!\{10^{i}\mid i\in\mathbb{Z}, -3\leq i\leq 0\}$ is conducted on the validation sets of the four datasets to assess performance sensitivity. Owing to larger node count and denser graph topology in Reddit-CL, we forgo $\alpha=64$ to avert out-of-memory (OOM) errors. Fig.~\ref{fig:params} presents the evolution of ADR performance under varying $\alpha$ and $\gamma$ settings. We observe consistent patterns across CoraFull-CL, Arxiv-CL, and Reddit-CL, with ADR performance improving as $\alpha$ increases. The influence of $\gamma$ is modest on the latter two datasets, whereas CoraFull-CL demands stronger regularization to preclude ill-conditioned LS solutions. On CS-CL, the trend is inverted: ADR attains superior performance with smaller $\alpha$. At $\alpha=1$, the feature buffer layer $\mathcal{B}(\cdot)$ can be considered removed. This is because the 128-dimensional feature space already ensures linear discriminability of classes in CS-CL, and further expansion risks introducing superfluous noise. The effect of $\gamma$ is likewise marginal at smaller $\alpha$.

\subsection{Profiling Severe Class Imbalance in Arxiv-CL}
\label{appendix:d}

\begin{figure*}[t]
  \begin{center}
    \centerline{\includegraphics[width=1.0\textwidth, trim=0 0 0 0, clip]{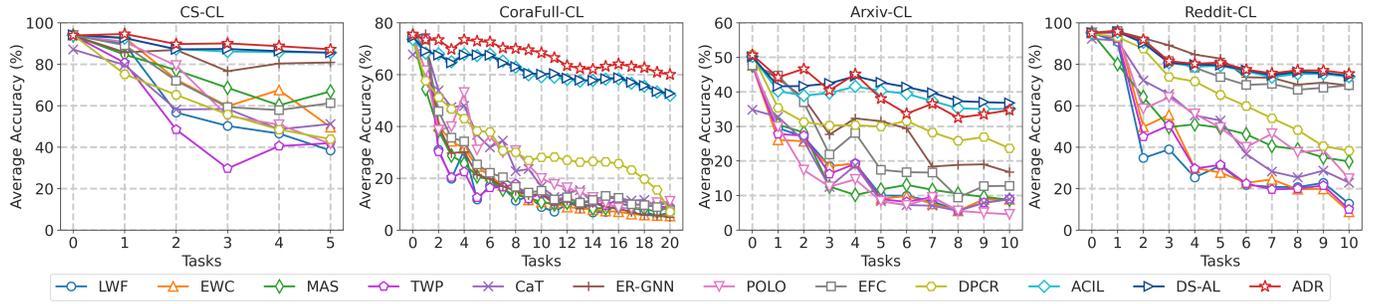}}
    \caption{
    Learning dynamics over the task streams on CS-CL, CoraFull-CL, Arxiv-CL, and Reddit-CL, with $\mathcal{A}_t$ reported for all tasks.
    }
    \label{fig_learningcurve}
  \end{center}
\end{figure*}

\begin{figure*}[t]
  \begin{center}
    \subfloat{
        \includegraphics[width=0.232\textwidth, trim=6 10 8 7, clip]{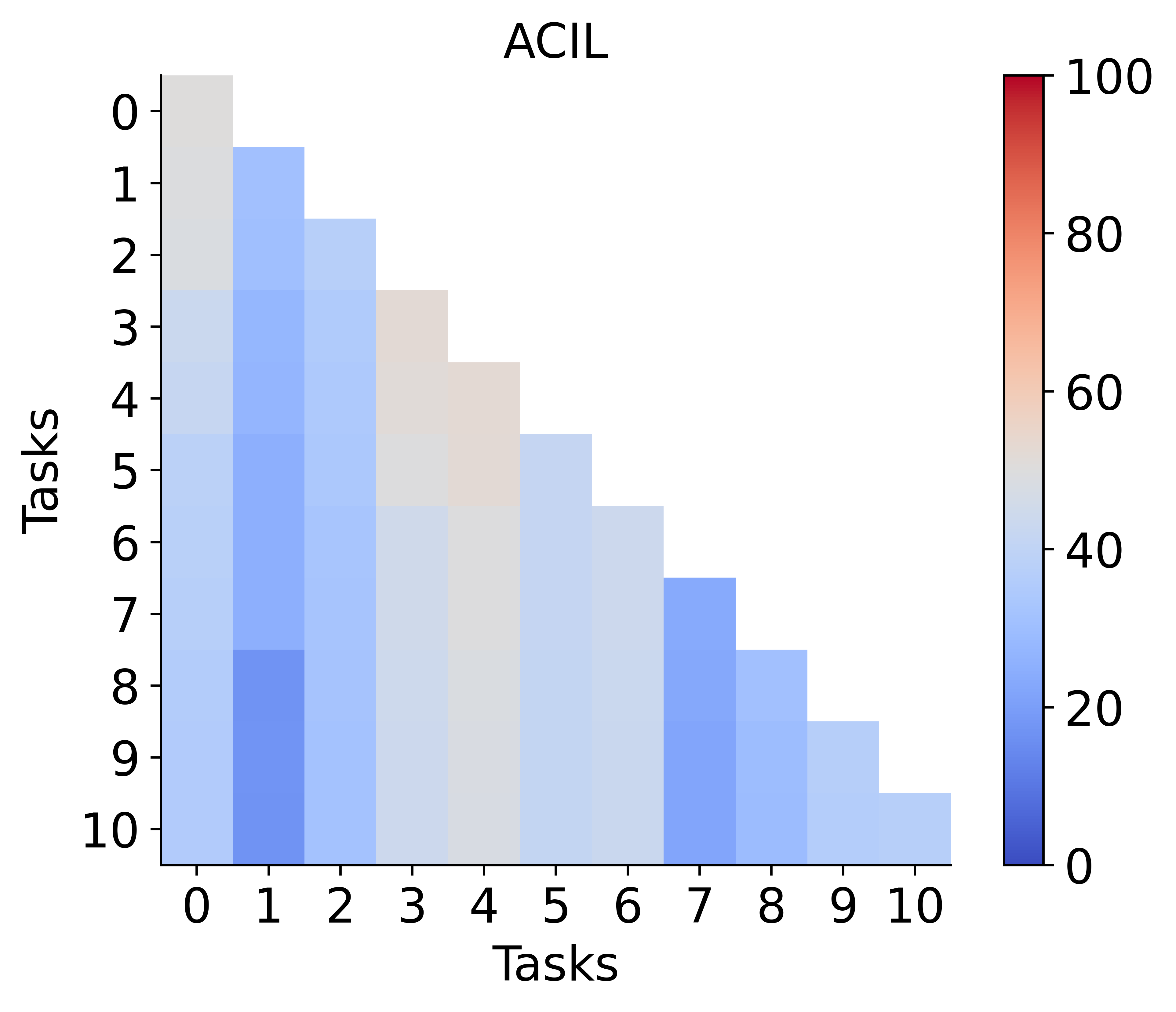}
        \label{fig:pm1}
    }
    \hfil
    \subfloat{
        \includegraphics[width=0.232\textwidth, trim=6 10 8 7, clip]{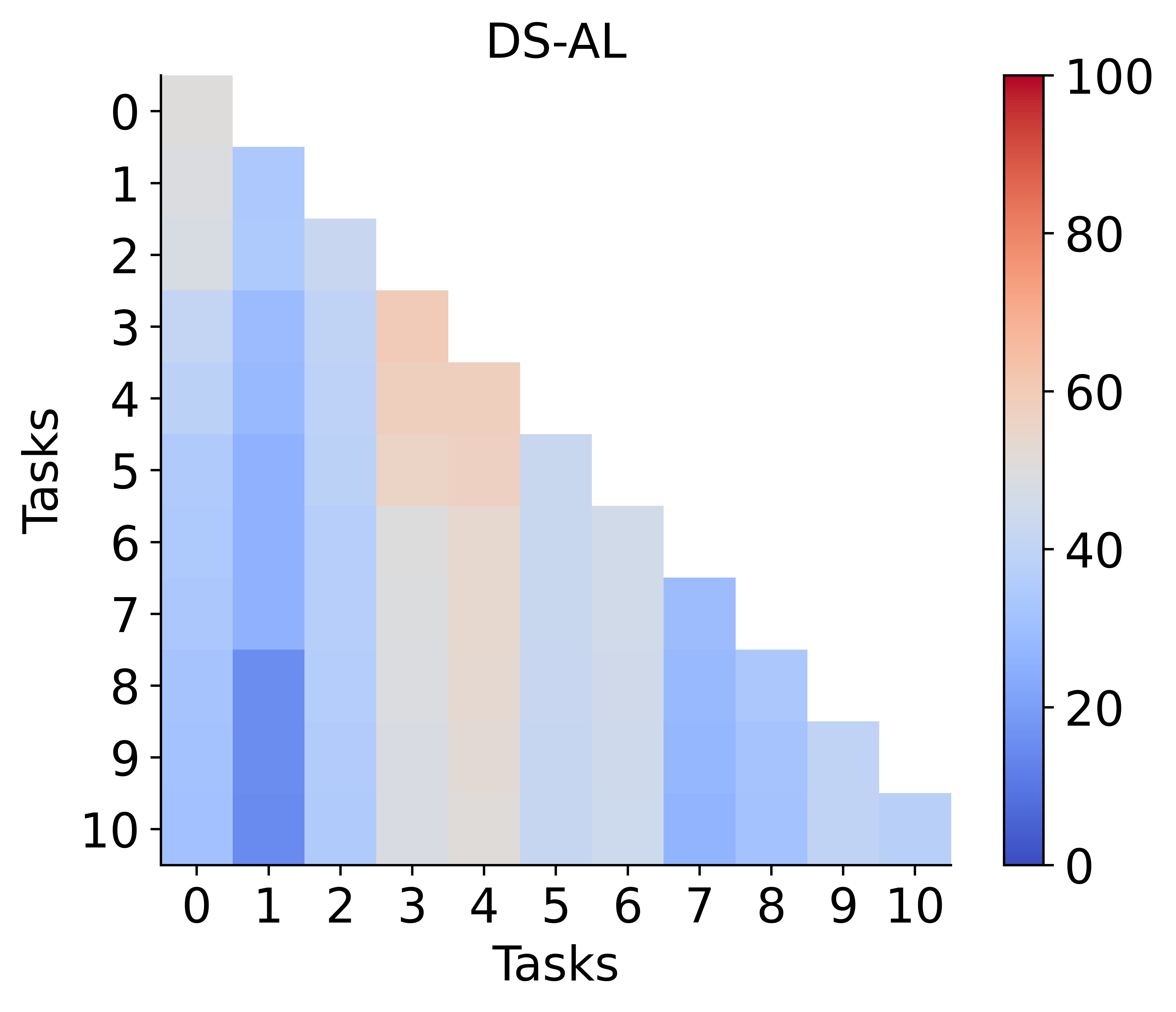}
        \label{fig:pm2}
    }
    \hfil
    \subfloat{
        \includegraphics[width=0.232\textwidth, trim=6 10 8 7, clip]{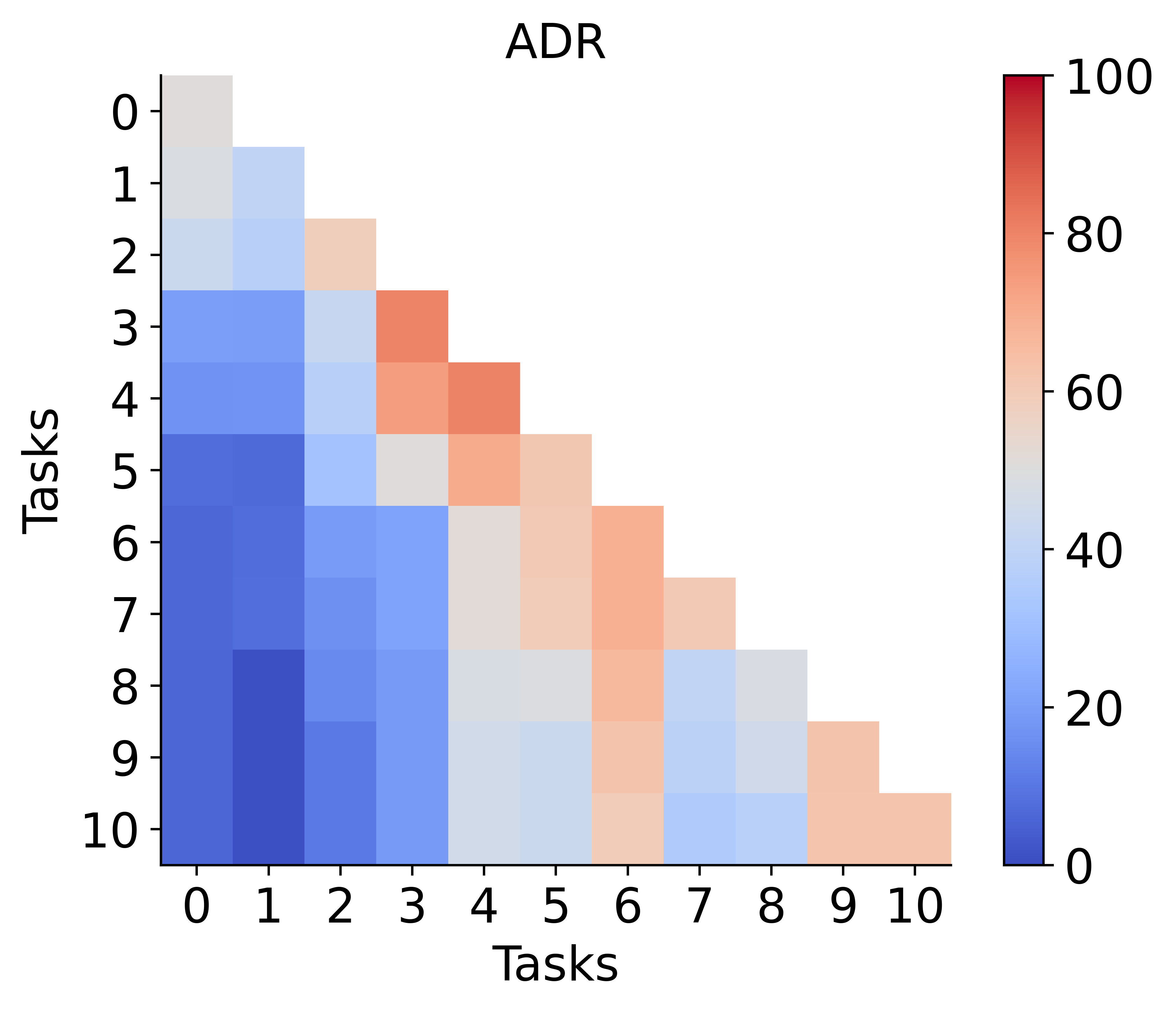}
        \label{fig:pm3}
    }
    \hfil
    \subfloat{
        \includegraphics[width=0.232\textwidth, trim=23 24 22 21, clip]{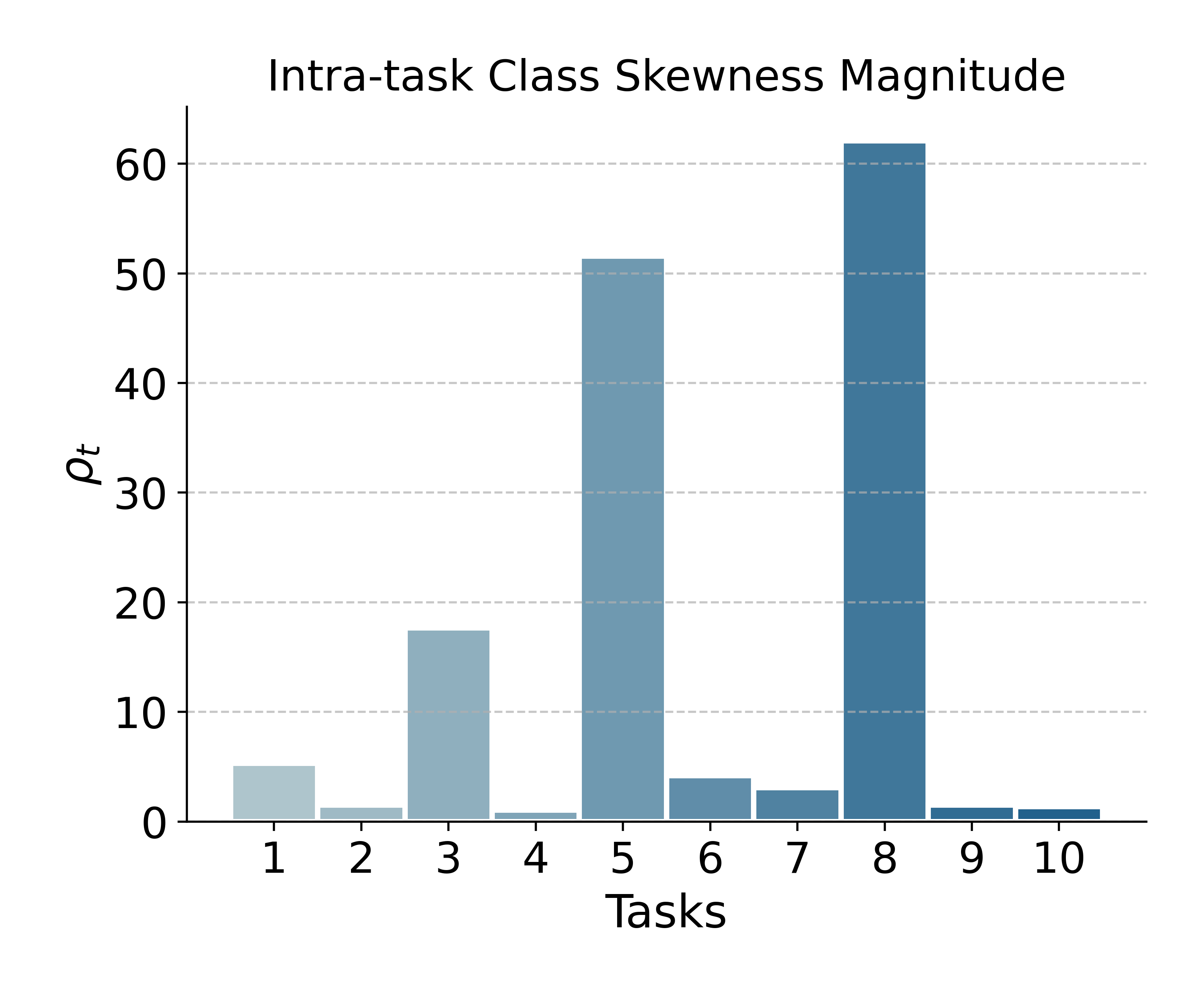}
        \label{fig:ratio}
    }
    \caption{Left three panels: Performance matrices for ACIL, DS-AL, and ADR on the test set of Arxiv-CL. Rightmost panel: Quantification of intra-task class imbalance in the training set of Arxiv-CL.}
    \label{fig:pm}
  \end{center}
\end{figure*}

As observed in Section~\ref{sub:Comparison}, the proposed ADR delivers notable performance improvements on CS-CL, CoraFull-CL, and Reddit-CL, while performing marginally below ACIL and DS-AL on Arxiv-CL overall. To probe the potential causes, we visualize the learning dynamics of all methods across the four benchmarks, as illustrated in Fig.~\ref{fig_learningcurve}. With respect to the Arxiv-CL benchmark, ADR outperforms ACIL and DS-AL on tasks $\mathcal{T}_{0:2}$ but undergoes a sharp trend inversion on tasks $\mathcal{T}_3$, $\mathcal{T}_5$, and $\mathcal{T}_8$. To elucidate this phenomenon, we further present the performance matrices of the three methods on Arxiv-CL (left three panels in Fig.~\ref{fig:pm}) and quantify the intra-task class imbalance for each incremental task $\mathcal{T}_{t>0}$ (rightmost panel in Fig.~\ref{fig:pm}). The intra-task class skewness magnitude $\rho_t$, serving as a measure of class imbalance, is formally defined as follows:
\begin{equation}
\label{eq:imbalance}
\rho_t=\frac{\max_{y_t\in\mathcal{Y}_t}\left|\mathcal{V}_{t}^{y_t}\right|}{\min_{y_t\in\mathcal{Y}_t}\left|\mathcal{V}_{t}^{y_t}\right|},
\end{equation}
where $\mathcal{V}_{t}^{y_t}=\{v_o\in\mathcal{V}_t\mid \mathrm{label}(v_o)=y_t\}$. Larger $\rho_t$ signifies greater severity of class imbalance. We observe that tasks $\mathcal{T}_3$, $\mathcal{T}_5$, and $\mathcal{T}_8$ manifest extreme class imbalance, which engenders pronounced bias in the trained GNNs and compromises ACL performance. This finding supports the pattern presented in Fig.~\ref{fig_learningcurve}. Moreover, the performance matrices in Fig.~\ref{fig:pm} reveal that following training on tasks $\mathcal{T}_3$, $\mathcal{T}_5$, and $\mathcal{T}_8$, all three methods exhibit a marked decline in test performance on tasks $\mathcal{T}_{0:2}$, as evidenced by abrupt changes in color. Compared with ACIL and DS-AL, ADR suffers more severe degradation, as the former freeze the encoder and expose only the classifier to class imbalance, whereas ADR updates the entire model, with both the encoder and the classifier being affected. A previous study~\cite{hammoud2024model} has shown that a biased individual model can induce bias in the merged model, even when the remaining individual models are unbiased. As a result, extreme class imbalance poses a challenge to the effectiveness of the proposed HAM. However, since class imbalance is not the primary focus of this paper, we defer its exploration to future work.

\section{Conclusion and Future Work}

In this paper, we proposed ADR, a theoretically grounded NECGL framework aimed at resolving compromised model plasticity and feature drift. We liberated the encoder from frozen parameters, enabling it to flexibly adapt to emerging task graph distributions. To circumvent feature drift induced by BP, we proposed HAM, which consolidates the parameters of GNNs at each layer by optimizing a joint ridge regression objective, conferring absolute resistance to feature drift. On this basis, we implemented ACR atop the merged encoder, which achieved theoretically zero-forgetting class-incremental learning by solving an analogous joint ridge regression problem. ADR demonstrated competitive performance compared with existing SOTA methods across four node classification benchmarks while strictly adhering to data privacy requirements. In the future, we will extend the proposed ADR framework to broader application scenarios---such as streaming recommendation, continual disease classification, and continual anomaly detection---while addressing class imbalance to improve its applicability under extreme conditions.


\bibliography{ref}
\bibliographystyle{IEEEtran}

\end{document}